\documentclass{article}

\PassOptionsToPackage{numbers, sort, compress}{natbib}

\usepackage[final]{neurips_2023}

\usepackage[utf8]{inputenc} 
\usepackage[T1]{fontenc}    
\usepackage[colorlinks=true, citecolor=blue, linkcolor=black, urlcolor=Violet]{hyperref}       
\usepackage{url}            
\usepackage{booktabs}       
\usepackage{amsfonts}       
\usepackage{nicefrac}       
\usepackage{microtype}      
\usepackage[dvipsnames]{xcolor}         
\usepackage{subfigure}
\usepackage{multirow}
\usepackage{multicol} 
\usepackage{pifont}
\usepackage{graphicx}
\usepackage{algorithm}
\usepackage{algorithmic}
\usepackage{caption}
\usepackage{amsmath}
\usepackage{amssymb}
\usepackage[font={footnotesize}]{caption}

\usepackage[utf8]{inputenc} 
\usepackage[T1]{fontenc}    
\usepackage{hyperref}       
\usepackage{url}            
\usepackage{booktabs}       
\usepackage{amsfonts}       
\usepackage{nicefrac}       
\usepackage{microtype}      

 \usepackage{graphicx}
\usepackage{subfigure} 
\usepackage{pifont}
\usepackage{adjustbox}
\usepackage{lipsum}
\usepackage{color}
\usepackage{wrapfig}
\usepackage{booktabs}
\usepackage[textsize=scriptsize]{todonotes}
\usepackage{multirow,mathtools}
\usepackage{threeparttable}

\definecolor{lightblue}{rgb}{0.68, 0.85, 0.9}
\definecolor{lightgreen}{rgb}{0.56, 0.93, 0.56}
\definecolor{lightskyblue}{rgb}{0.53, 0.81, 0.98}
\definecolor{non-photoblue}{rgb}{0.64, 0.87, 0.93}
\definecolor{magicmint}{rgb}{0.67, 0.94, 0.82}
\definecolor{mossgreen}{rgb}{0.68, 0.87, 0.68}
\definecolor{salmon}{rgb}{1.0, 0.55, 0.41}
\definecolor{babypink}{rgb}{0.96, 0.76, 0.76}
\definecolor{darkgreen}{rgb}{0, 0.7, 0}

\newtheorem{myprop}{\bf{Proposition}}

\DeclareMathAlphabet\mathbfcal{OMS}{cmsy}{b}{n}
\newcommand{\Def}[0]{\mathrel{\mathop:}=}

 \newcommand{\YL}[1]{\textcolor{orange}{YL: #1}}
 
 \newcommand{\SL}[1]{\textcolor{purple}{SL: #1}}
 \newcommand{\JH}[1]{\textcolor{blue}{JH: #1}}
 \newcommand{\JC}[1]{\textcolor{darkgreen}{JC: #1}}

\def\btheta{\boldsymbol{\theta}}

\usepackage{color, colortbl}
\definecolor{Gray}{gray}{0.93}
\definecolor{Orange}{rgb}{1,0.5,0}
\definecolor{DGray}{gray}{0.83}
\definecolor{LightCyan}{rgb}{0.88,1,1}

\usepackage[most]{tcolorbox}
\newtcolorbox{mybox}[2][]{%
  attach boxed title to top center
               = {yshift=-8pt},
  colback      = Gray,
  colframe     = black,
  fonttitle    = \bfseries,
  colbacktitle = white,
  title        = #2,#1,
  enhanced,
}

\DeclarePairedDelimiterX{\inp}[2]{\langle}{\rangle}{#1, #2}

\DeclareMathOperator*{\argmin}{arg\,min}

\makeatletter
\newcommand*{\rom}[1]{\expandafter\@slowromancap\romannumeral #1@}
\makeatother
\newcommand{\mycomment}[1]{}

\definecolor{Sijia_color}{rgb}{0.858, 0.188, 0.478}

\newcommand{\MU}{{\text{MU}}}
\newcommand{\Df}{\mathcal D_\mathrm{f}}
\newcommand{\Dr}{\mathcal D_\mathrm{r}}
\newcommand{\thetaunl}{\boldsymbol \theta_\mathrm{u}}
\newcommand{\thetafull}{\boldsymbol \theta_\mathrm{o}}
\newcommand{\Lunl}{L_\mathrm{u}}

\newcommand{\retrain}{{\text{Retrain}}}
\newcommand{\FT}{{\text{FT}}}
\newcommand{\GA}{{\text{GA}}}
\newcommand{\FF}{{\text{FF}}}
\newcommand{\IU}{{\text{IU}}}

\newcommand{\retrainC}{{\color{red}{\text{Retrain}}}}
\newcommand{\FTC}{{\color{ForestGreen}{\text{FT}}}}
\newcommand{\GAC}{{\color{blue}{\text{GA}}}}
\newcommand{\FFC}{{\color{YellowOrange}{\text{FF}}}}
\newcommand{\IUC}{{\color{purple}{\text{IU}}}}

\newcommand{\UA}{{\text{UA}}}
\newcommand{\RA}{{\text{RA}}}
\newcommand{\TA}{{\text{TA}}}

\newcommand{\MIAF}{{\text{MIA}-Efficacy}}
\newcommand{\MIAR}{{\text{MIA}-Privacy}}
\newcommand{\RTE}{{\text{RTE}}}

\newcommand{\MUSparse}{{\text{$\ell_1$-sparse MU}}}

\newcommand{\acc}{{\text{Acc}}}
\newcommand{\TIME}{{\text{Time}}}

\title{
Model Sparsity Can Simplify Machine Unlearning
}

\author{%
  Jinghan Jia$^{1, \star}$
  \And Jiancheng Liu$^{1, \star}$
  \And Parikshit Ram$^{2}$ 
  \And Yuguang Yao$^{1}$
  \And Gaowen Liu$^{3}$
  \And Yang Liu$^{4,5}$
  \And Pranay Sharma$^{6}$
  \And Sijia Liu$^{1,2}$   
  \AND \vspace*{-5mm}\\
  ${}^1$Michigan State University,
  ${}^2$IBM Research,
  ${}^3$Cisco Research,\\
  ${}^4$University of California, Santa Cruz,
   ${}^5$ByteDance Research,
  ${}^6$Carnegie Mellon University\\
  $^\star$Equal contribution\\
}

\begin{document}

\maketitle


\begin{abstract}
In response to recent data regulation requirements, machine unlearning (\MU) has emerged as a critical process to remove the influence of specific examples from a given model. Although {exact} unlearning can be achieved through complete model retraining using the remaining dataset, the associated computational costs have driven the development of efficient, approximate unlearning techniques. Moving beyond data-centric MU approaches, our study introduces a novel model-based perspective: model sparsification via weight pruning, which is capable of reducing the gap between exact unlearning and approximate unlearning. 
We show in both theory and practice  that model sparsity can boost the multi-criteria unlearning performance of an approximate unlearner, closing the approximation gap, while continuing to be efficient. This   leads to a new MU paradigm,    termed {prune first, then unlearn}, which infuses a sparse model prior into the unlearning process. Building on this insight, we  also develop a sparsity-aware unlearning method that utilizes sparsity regularization to enhance the training process of   approximate unlearning.
Extensive experiments show that our  proposals consistently benefit MU in various unlearning scenarios. A notable highlight is the 77\% unlearning efficacy gain of fine-tuning (one of the simplest  unlearning methods) when    using sparsity-aware unlearning.
Furthermore, we  demonstrate the practical impact of  our proposed MU methods in addressing  other machine learning challenges, such as defending against backdoor attacks and  enhancing transfer learning.
{Codes are available at \url{https://github.com/OPTML-Group/Unlearn-Sparse}.}

\end{abstract}
\section{Introduction}

Machine unlearning (\textbf{\MU}) initiates a reverse learning process to scrub the influence of   data points   from a trained machine learning (\textbf{ML}) model. It was  introduced to avoid information leakage about private   data upon completion of training \cite{cao2015towards, bourtoule2021machine, nguyen2022survey},
particularly in compliance with
legislation like `the right to be forgotten' \cite{rosen2011right} in General Data Protection Regulation (GDPR) \cite{hoofnagle2019european}.
The  \textit{direct but optimal} unlearning approach  is \textit{exact unlearning} to \textit{retrain}   ML models from scratch using the  remaining training set, after removing the data points to be scrubbed.
Although retraining yields the \textit{ground-truth} unlearning strategy, it is the most computationally intensive one.
Therefore,   the development of \textit{approximate but fast} unlearning methods   has  become a  major focus in research \cite{warnecke2021machine,graves2021amnesiac,thudi2021unrolling,becker2022evaluating,izzo2021approximate}.

Despite the computational benefits of  approximate unlearning, it often  lacks a strong guarantee on the effectiveness  of unlearning, resulting in a performance gap with  exact unlearning \cite{thudi2022necessity}.  In particular, we encounter two main challenges.
\textit{First}, the performance of approximate unlearning   can heavily rely on the configuration of algorithmic parameters.
For example, the Fisher forgetting method \cite{golatkar2020eternal} needs to carefully tune the Fisher information regularization parameter  in each   data-model setup.  
\textit{Second}, the effectiveness of an approximate scheme can vary significantly across the different unlearning evaluation criteria, and their trade-offs are not well understood. For example,  high `efficacy' (ability to protect the privacy of the scrubbed data)
\textit{neither} implies \textit{nor} precludes high `fidelity' (accuracy on the remaining dataset) \cite{becker2022evaluating}. This raises our \textbf{driving question (Q)}  below:
\vspace*{3mm}
\begin{tcolorbox}[before skip=-2.1mm, after skip=0.2cm, boxsep=0.0cm, middle=0.1cm, top=0.1cm, bottom=0.1cm]
\textbf{(Q)} \textit{Is there a theoretically-grounded and broadly-applicable method to improve approximate unlearning across different unlearning criteria?}
\end{tcolorbox}

\begin{wrapfigure}{r}{80mm}
\vspace*{-4mm}
\centerline{
\includegraphics[width=80mm,height=!]{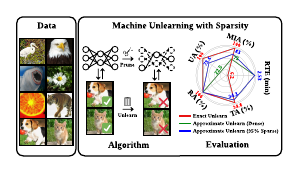}
}
\vspace*{-2mm}
\caption{\footnotesize{
Schematic overview of our proposal on model sparsity-driven {\MU}. Evaluation at-a-glance shows the performance of three unlearning methods (retraining-based exact unlearning, finetuning-based approximate unlearning \cite{golatkar2020eternal}, 
and proposed   unlearning on 95\%-sparse model) under five metrics:  
unlearning accuracy ({\UA}), membership inference attack (MIA)-based unlearning efficacy, accuracy on remaining data ({\RA}), testing accuracy (\TA), and run-time efficiency (\RTE); see summary in \textbf{Tab.\,\ref{tab: summary_MU_methods_metrics}}.
The unlearning scenario is given by class-wise forgetting, where data points of a single class are scrubbed.  Each metric is normalized to $[0,1]$ based on the best result across unlearning methods 
  for ease of visualization.
Results indicate that \textcolor{blue}{model sparsity} reduces the gap between \textcolor{red}{exact} and \textcolor{ForestGreen}{approximate} {\MU} without loss in efficiency. 
}}
\label{fig: results_highlights}
\vspace{-7.1mm}
\end{wrapfigure}

To address \textbf{(Q)}, we advance {\MU} through a fresh and novel viewpoint: \textbf{model sparsification}. \textit{Our key finding} is that model sparsity (achieved by weight pruning) can significantly reduce the gap between approximate unlearning and exact unlearning; see \textbf{Fig.\,\ref{fig: results_highlights}} for the schematic overview of our proposal
and  highlighted empirical performance.

Model sparsification (or weight pruning) has been extensively studied   in the literature \cite{han2015deep, chen2021lottery, frankle2018lottery, frankle2020linear, ma2021sanity, zhang2022advancing, blalock2020state}, 
focusing on the interrelation between model compression and generalization.
For example, the notable lottery ticket hypothesis (\textbf{LTH})  \cite{frankle2018lottery} demonstrated 
the existence of a sparse subnetwork (the so-called `winning ticket') that matches or even exceeds the test accuracy of the original dense model. In addition to generalization, the impact of pruning has also been investigated on model robustness \cite{sehwag2020hydra,chen2022quarantine,diffenderfer2021winning}, fairness \cite{stoychev2022effect,xu2022can}, interpretability \cite{wong2021leveraging,chen2022can}, loss landscape \cite{frankle2020linear,chen2022can}, and privacy \cite{huang2020privacy,wang2020against}. In particular, 
the privacy gains from pruning \cite{huang2020privacy,wang2020against} imply connections between data influence and model sparsification.


More recently, 
a few  works   \cite{wang2022federated,ye2022learning} attempted to draw insights from pruning for unlearning. In \citet{wang2022federated},  removing channels of a  deep neural network (\textbf{DNN}) showed an  unlearning benefit in federated learning. And in \citet{ye2022learning}, filter pruning   was introduced in lifelong learning to detect ``pruning identified exemplars'' \cite{hooker2019compressed} that  are easy to forget.
\textbf{However}, different from the above literature that    customized  model pruning   for a specific unlearning application,  our work systematically and comprehensively explores and exploits the foundational connections between unlearning and pruning.
We summarize our \textbf{contributions} below.

$\bullet$ First, we provide  a holistic understanding of  {\MU} across the full training/evaluation stack. 

$\bullet$ Second, we draw a tight connection between {\MU} and model pruning and show in   theory and practice that model sparsity   helps  close the gap between approximate unlearning and exact unlearning.

$\bullet$ Third, we develop a new {\MU} paradigm termed `prune first, then unlearn', and investigate the influence of pruning methods in the performance of unlearning. Additionally, we   develop a novel `sparsity-aware unlearning' framework that leverages a soft sparsity regularization scheme to enhance the approximate unlearning process.

$\bullet$ Finally, we perform extensive experiments across diverse datasets, models, and unlearning scenarios. Our findings consistently highlight the crucial role of model sparsity in enhancing \MU.





\section{Revisiting Machine Unlearning and Evaluation}
\label{sec: primer_MU}

\noindent \textbf{Problem setup.}
MU aims to remove ({or} scrub) the influence of some targeted training data on a trained ML model \cite{cao2015towards,bourtoule2021machine}. 
Let $\mathcal{D} = \{\mathbf z_i \}_{i=1}^N$ be a (training) dataset of $N$  data points, with label information encoded for supervised learning. $\Df \subseteq \mathcal D$ represents  a subset whose influence we want to scrub, termed
the \textbf{forgetting dataset}.
Accordingly, the complement of $\Df$ is the \textbf{remaining dataset}, \textit{i.e.}, $\Dr = \mathcal D \setminus \mathcal{D}_{\mathrm{f}}$.
We denote by $\btheta$ the model parameters, and $\thetafull$   the \textbf{original  model} trained    on the entire training set  $\mathcal D$ using \textit{e.g.}, empirical risk minimization (ERM). Similarly, we denote by $\thetaunl$  an \textbf{unlearned model}, obtained by a scrubbing algorithm, after    removing the influence of $\Df$ from the  trained model $\thetafull$. The \textbf{problem of MU} is to find an accurate and efficient  scrubbing mechanism  to generate  $\thetaunl$   from  $\thetafull$. 
In existing studies \cite{golatkar2020eternal,graves2021amnesiac,bourtoule2021machine}, the choice of the forgetting dataset $\Df$ specifies different unlearning scenarios. 
There exist two main categories. 
{First}, \textit{class-wise forgetting} \cite{golatkar2020eternal,graves2021amnesiac} refers to  unlearning $\Df$ consisting of training data points of an entire class. 
Second, \textit{random data forgetting} corresponds to unlearning $\Df$  given by a  subset of random data drawn from all classes.




%
%

\noindent \textbf{Exact and approximate MU methods.} The \textit{exact unlearning}  method refers to 
\textit{retraining} the model parameters  from \textit{scratch} over the remaining dataset $\Dr$. 
Although retraining from scratch (that we term  \textbf{\retrain})
is optimal for MU, it entails a large computational overhead, particularly for DNN training. This problem is alleviated by \textit{approximate unlearning}, an easy-to-compute proxy for {\retrain}, which has received growing attention.
Yet,   the boosted computation efficiency   comes at the cost of  MU's  efficacy. 
We next review some commonly-used approximate unlearning methods that we improve in the sequel by leveraging sparsity; see a summary in  \textbf{Tab.\,\ref{tab: summary_MU_methods_metrics}}. 




\begin{wraptable}{r}{70mm}
\vspace*{-4.5mm}
        \centering
         \caption{\footnotesize{Summary of approximate unlearning methods considered in this work. 
         The marker `\textcolor{black}{\ding{51}}' denotes the metric used in previous research. The number in {\RTE} is the   run-time cost reduction compared to the  cost of {\retrain}, based on our empirical studies in Sec.\,\ref{sec: exp} on 
         (CIFAR-10, ResNet-18). Note that 
         {\GA} seems better than ours in terms of RTE, but it is less effective in unlearning.
         } 
         }
\vspace*{-2mm}
        \label{tab: summary_MU_methods_metrics}
        \resizebox{0.50\textwidth}{!}{
        \begin{tabular}{c|ccccc|c}
        \toprule
        Unlearning &  \multicolumn{5}{c|}{Evaluation metrics} & \multirow{2}{*}{Representative work}\\
        Methods &  \UA& \MIAF & \RA & \TA & \RTE & \\
        \midrule
        FT & \textcolor{black}{\ding{51}} & & \textcolor{black}{\ding{51}}& \textcolor{black}{\ding{51}}& 0.06$\times$ & \cite{golatkar2020eternal,warnecke2021machine}    \\
        GA &  \textcolor{black}{\ding{51}}& \textcolor{black}{\ding{51}}& \textcolor{black}{\ding{51}}& \textcolor{black}{\ding{51}}& 0.02$\times$ &  \cite{graves2021amnesiac,thudi2021unrolling}  \\
        FF & \textcolor{black}{\ding{51}} & & \textcolor{black}{\ding{51}}& \textcolor{black}{\ding{51}}& 0.9
        $\times$ & \cite{golatkar2020eternal,becker2022evaluating}    \\
        \IU & \textcolor{black}{\ding{51}} & & &\textcolor{black}{\ding{51}} & 0.08$\times$ &  \cite{koh2017understanding,izzo2021approximate} \\
        \midrule
        Ours & \textcolor{black}{\ding{51}} & \textcolor{black}{\ding{51}}&\textcolor{black}{\ding{51}} & \textcolor{black}{\ding{51}}& 0.07$\times$ & This work  \\
        \bottomrule
        \end{tabular}}
\vspace*{-7.5mm}
\end{wraptable}
\noindent\ding{70} \textit{Fine-tuning (\textbf{\FT})} \cite{golatkar2020eternal,warnecke2021machine}:  
Different from    {\retrain}, {\FT} fine-tunes the   pre-trained  model   $\thetafull$ on {$\Dr$} using a few training epochs to obtain ${\thetaunl}$. The rationale is that fine-tuning on {$\Dr$} initiates the catastrophic forgetting in  the model over {$\Df$} as is common in continual learning \cite{parisi2019continual}. 

\noindent\ding{70} \textit{Gradient ascent (\textbf{\GA})} \cite{graves2021amnesiac,thudi2021unrolling}:
{\GA} reverses the model training on  
$\Df$ by adding the corresponding gradients    back to  $\thetafull$, \textit{i.e.}, moving $\thetafull$  in the direction of increasing  loss for   data points to be scrubbed. 


\noindent \ding{70} \textit{Fisher forgetting (\textbf{\FF})}  \cite{golatkar2020eternal,becker2022evaluating}:
{\FF} adopts  an  additive Gaussian noise  to `perturb'   $\thetafull$ towards   exact unlearning.
Here the Gaussian distribution has zero mean  and covariance determined by  the $4$th root of Fisher Information matrix with respect to (w.r.t.) $\thetafull$ on   $\Dr$.
We note that the computation of the Fisher Information matrix exhibits lower parallel efficiency in contrast to other unlearning methods, resulting in higher computational time when executed on GPUs; see
\citet{golatkar2020eternal} for implementation details. 

\noindent \ding{70} \textit{Influence unlearning (\textbf{\IU})} \cite{koh2017understanding,izzo2021approximate}:
{\IU} leverages the influence function approach \cite{cook1982residuals} to characterize the change in  $\thetafull$  if a training point is removed from the training loss. {\IU} estimates the change in model parameters from $\thetafull$ to $\thetaunl$, \textit{i.e.}, $\thetaunl - \thetafull$.  {{\IU} also relates to an important line of research in MU, known as $\epsilon$-$\delta$ forgetting \cite{wang2022federated,guo2019certified,xu2023machine}. However, it typically requires additional model and training assumptions \cite{guo2019certified}. 
}

We next take a step further to revisit the {\IU} method and re-derive its formula (\textbf{Prop.\,\ref{prop: IU}}), with the aim of enhancing the effectiveness of existing  solutions proposed in the previous research.

\begin{myprop}
\label{prop: IU}
Given the weighted ERM training $\btheta(\mathbf w) = \argmin_{\btheta}
L(\mathbf w,\btheta)
$ where $L(\mathbf w,\btheta) = \sum_{i=1}^N [w_i \ell_i (\btheta, \mathbf z_i)]$, 
$w_i \in [0,1]$ is the influence weight associated with the data point $\mathbf z_i$ and $\mathbf 1^T \mathbf w = 1$, 
the  model update from   $\thetafull$  to $\btheta(\mathbf w)$ yields 
{\small\begin{align}
   \hspace*{-2mm} \Delta(\mathbf w) \Def   \btheta(\mathbf w) - \thetafull \approx  \mathbf H^{-1} {\color{black}{  \nabla_{\btheta} L( \mathbf 1/N - \mathbf w, \thetafull)  }},  
  \hspace*{-2mm}
  \label{eq: Delta_IU}
\end{align}}%
where $\mathbf 1$ is the $N$-dimensional vector of all ones, {$\mathbf w=\mathbf{1}/N$ signifies the uniform weights used by ERM}, $\mathbf H^{-1} $ is the inverse of the Hessian    $\nabla^2_{\btheta,\btheta} L(\mathbf 1/N,  \thetafull)$ evaluated at $ \thetafull$, and $\nabla_{\btheta} L$ is the gradient of $L$. When scrubbing $\Df$, the unlearned model is given by
$\thetaunl = \thetafull + \Delta(\mathbf{w}_\mathrm{MU})$. Here  
$\mathbf w_{\mathrm{MU}} \in [0, 1]^{N}$ with entries $w_{\mathrm{MU},i} = \mathbb{I}_{\Dr } (i) / |\Dr|$ signifying the data influence weights for {\MU}, 
$\mathbb{I}_{\Dr } (i) $ is the indicator function with value 1 if $i \in \Dr $ and 0 otherwise, and $|\Dr|$ is the cardinality of $\Dr$.
\end{myprop}
%
\textbf{Proof}: We derive \eqref{eq: Delta_IU}  using an implicit gradient approach;
see Appendix\,\ref{appendix: IU}.
\hfill $\square$


It is worth noting that we have taken into consideration the weight normalization effect $\mathbf 1^T \mathbf w =1$ 
 in  \eqref{eq: Delta_IU}. This is   different from  existing work like \citet[Sec.\,3]{izzo2021approximate} using Boolean or unbounded weights. In practice, we found that {\IU} with  weight normalization can improve the unlearning performance. 
Furthermore, 
to update the model influence given by \eqref{eq: Delta_IU}, one needs to acquire the second-order information in the form of  inverse-Hessian gradient product. Yet, the exact computation is prohibitively expensive. To overcome this issue, we use the first-order WoodFisher approximation \cite{singh2020woodfisher} to estimate  the inverse-Hessian gradient product. 

\noindent \textbf{Towards a `full-stack' {\MU} evaluation.}
Existing work   has assessed  {\MU} performance from different aspects \cite{thudi2021unrolling,golatkar2020eternal,graves2021amnesiac}. 
Yet, a single  performance metric may provide   a limited 
view of {\MU} \cite{thudi2022necessity}. 
By carefully reviewing the prior art, we focus on the following empirical metrics (summarized in  Tab.\,\ref{tab: summary_MU_methods_metrics}).

\noindent \ding{70} \textit{Unlearning accuracy (\textbf{\UA})}: 
We define $\mathrm{\UA}(\thetaunl) = 1-\mathrm{Acc}_{\Df} (\thetaunl)$ to characterize the \textit{efficacy} of {\MU} in the accuracy dimension, where $\mathrm{Acc}_{\Df} (\thetaunl)$ is the  accuracy  of $\thetaunl$ on the forgetting dataset  $\Df$ \cite{golatkar2020eternal,graves2021amnesiac}.
It is important to note that a more favorable {\UA}   for an approximate unlearning method should   \textbf{reduce its performance disparity  with  the gold-standard retrained model (\retrain)}; a higher value is not necessarily better. This principle also extends to other evaluation metrics.



\noindent \ding{70} \textit{Membership inference attack (MIA)  on  $\Df$ (\textbf{\MIAF})}:   This is another metric   to assess the \textit{efficacy} of unlearning.
It is achieved by applying the   confidence-based MIA predictor \cite{song2019privacy,yeom2018privacy} to the unlearned model ($\thetaunl$) on the forgetting dataset ($\Df$). The MIA success rate can then indicate how many samples in $\Df$ can be correctly predicted as forgetting (\textit{i.e.}, non-training) samples of $\thetaunl$.
 A \textit{higher} {\MIAF}   implies    less information about   $\Df$     in $\thetaunl$; see Appendix\,\ref{appendix: metric settings} for more details.

\noindent \ding{70} \textit{Remaining accuracy (\textbf{\RA})}:   This refers to 
the accuracy of {$\thetaunl$} on $\Dr$, which reflects the \textit{fidelity} of    {\MU} 
 \cite{becker2022evaluating}, {\textit{i.e.}, training data information}  should be  preserved from {$\thetafull$} to {$\thetaunl$}. 


\noindent \ding{70} \textit{Testing accuracy (\textbf{\TA})}:  This measures the \textit{generalization} ability of {$\thetaunl$} on  a testing dataset rather than $\Df$ and $\Dr$. {{\TA} is evaluated on the whole test dataset, except for class-wise forgetting, in which testing data points belonging to the forgetting class are not in the testing scope. 
}


\noindent \ding{70} \textit{Run-time efficiency (\textbf{\RTE})}:  This measures the computation efficiency of an {\MU} method.  For example,
if we regard the run-time cost of  {\retrain} as the baseline, the computation acceleration gained by different approximate unlearning methods is summarized in Tab.\,\ref{tab: summary_MU_methods_metrics}.

\section{
Model Sparsity: A Missing Factor Influencing Machine Unlearning
}
\label{sec: sparsityMU}



\begin{wrapfigure}{r}{45mm}
\vspace*{-5mm}
\centerline{
\includegraphics[width=45mm,height=!]{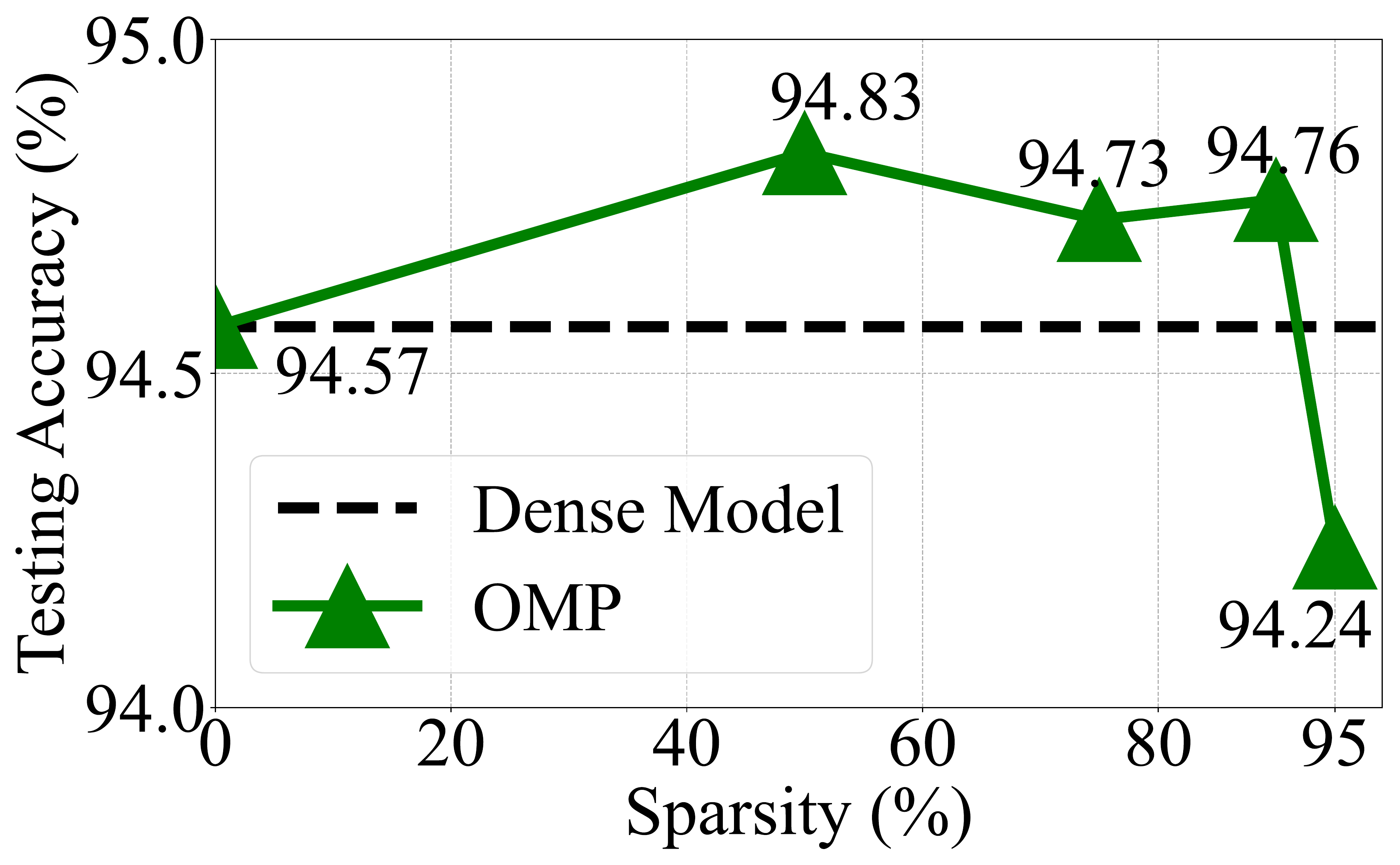}
}
\vspace*{-2mm}
\caption{\footnotesize{
{Testing accuracy  of   OMP-based sparse ResNet-18   vs. the   dense model on CIFAR-10. 
}
}}
  \label{fig: OMP_results}
 \vspace*{-5mm}
\end{wrapfigure}

\noindent \textbf{Model sparsification via weight pruning.}
Model sparsification could not only facilitate  a model's training, inference, and deployment  but also benefit model's performance. For example, LTH (lottery ticket hypothesis) \cite{frankle2018lottery} stated that a trainable sparse sub-model  could be identified from the original dense model, with test accuracy on par or even better than the original model. 
\textbf{Fig.\,\ref{fig: OMP_results}} shows an example of  the pruned model's generalization   vs. its   sparsity ratio. Here  one-shot magnitude pruning (\textbf{OMP}) \cite{ma2021sanity} is adopted to obtain sparse models. 
OMP is computationally the lightest pruning method, which   directly prunes the model weights to the target sparsity ratio based on their magnitudes. As we can see, there exists a graceful sparse regime  with lossless testing accuracy.

\noindent \textbf{Gains of {\MU} from sparsity.}
We first analyze the impact of model sparsity on {\MU}  through a lens of   \textit{unrolling stochastic gradient descent} (\textbf{SGD}) \cite{thudi2021unrolling}. The specified SGD method allows us to derive the \textit{unlearning error} (given by the  weight difference between the approximately unlearned model   and the
gold-standard retrained model)  when scrubbing a single data point. However, different from \citet{thudi2021unrolling}, we  will infuse the model sparsity into  SGD unrolling.

Let us assume a binary mask $\mathbf m$ associated with the model parameters $\btheta$, where $m_i = 0$ signifies that the $i$th parameter 
 $\theta_i$ is pruned to zero and $m_i = 1$  represents the unmasked   $\theta_i$.  This sparse pattern $\mathbf m$ could be obtained by a weight pruning method, like OMP. 
Given $\mathbf m$,   the \textbf{sparse model}  is $ \mathbf m \odot \btheta $, where $\odot$ denotes the element-wise multiplication. Thudi \emph{et al.}\,\cite{thudi2021unrolling} showed that if 
{\GA} is adopted to scrub a single data point for the original (dense) model $ \btheta$ (\textit{i.e.}, $\mathbf{m} = \mathbf 1$),  then  the gap between {\GA} and {\retrain} can be approximately bounded in the weight space. 
  \textbf{Prop.\,\ref{prop: SGD_sparse_MU}} extends the existing unlearning error analysis   to a    sparse model. 

\begin{myprop}
\label{prop: SGD_sparse_MU}
Given the model sparse pattern $\mathbf m$ and the SGD-based  training, the unlearning error   of  {\GA}, denoted by $e(\mathbf m)$, can be characterized by the weight distance
between the {\GA}-unlearned model and the gold-standard retrained model. This leads to the error bound
%
%

\vspace*{-5mm}
{\small\begin{align}
  e(\mathbf m) = \mathcal{O}(\eta^2 t  \| 
   \mathbf m \odot (\btheta_t - \btheta_0) \|_2 \sigma(\mathbf m) )
   \label{eq: err_bd_SGD_sparse}
\end{align}}%
where $\mathcal O$ is the big-O notation, $\eta$ is the learning rate, 
$t$ is the number of   training iterations, 
$ (\btheta_t - \btheta_0)$ denotes the weight difference at iteration $t$ from its   initialization $\btheta_0$,  and
$\sigma(\mathbf m)  $ is the largest singular value ($\sigma$) of the  Hessian $\nabla_{\btheta,\btheta}^2\ell$ (for a training loss $\ell$)   among the  unmasked parameter dimensions, \textit{i.e.}, $\sigma(\mathbf m) \Def \max_{j} \{  \sigma_{j}( \nabla_{\btheta,\btheta}^2\ell ), \text{if } m_j \neq 0  \}$.

\textbf{Proof}: 
See Appendix\,\ref{appendix: SGD_sparse_MU}. 
\hfill $\square$
\end{myprop}
We next draw some key insights   from \textbf{Prop.\,\ref{prop: SGD_sparse_MU}}. \textit{First},  it is clear from  \eqref{eq: err_bd_SGD_sparse} that the unlearning error reduces as the model sparsity in $\mathbf m$ increases. 
By contrast, the unlearning error  derived in  \citet{thudi2021unrolling} for a  dense model  (\textit{i.e.}, $\mathbf{m} = \mathbf 1$) is proportional to the dense model distance $\| \btheta_t - \btheta_0 \|_2$. 
Thus,   model sparsity  is beneficial to reducing the gap between ({\GA}-based) approximate and exact unlearning. 
\textit{Second}, the error bound \eqref{eq: err_bd_SGD_sparse}  enables us to relate {\MU} to the  spectrum of the  Hessian  of the loss landscape. 
The number of active singular values (corresponding to nonzero dimensions in $\mathbf m$) decreases when the sparsity grows.
However, it is important to note that in a high-sparsity regime, the model's generalization   could   decrease. Consequently, it is crucial to  select the model sparsity to strike a balance between generalization   and unlearning performance.

\begin{figure*}[t]
\centerline{
\begin{tabular}{cccc}
    \hspace*{-2mm}  \includegraphics[width=0.25\textwidth,height=!]{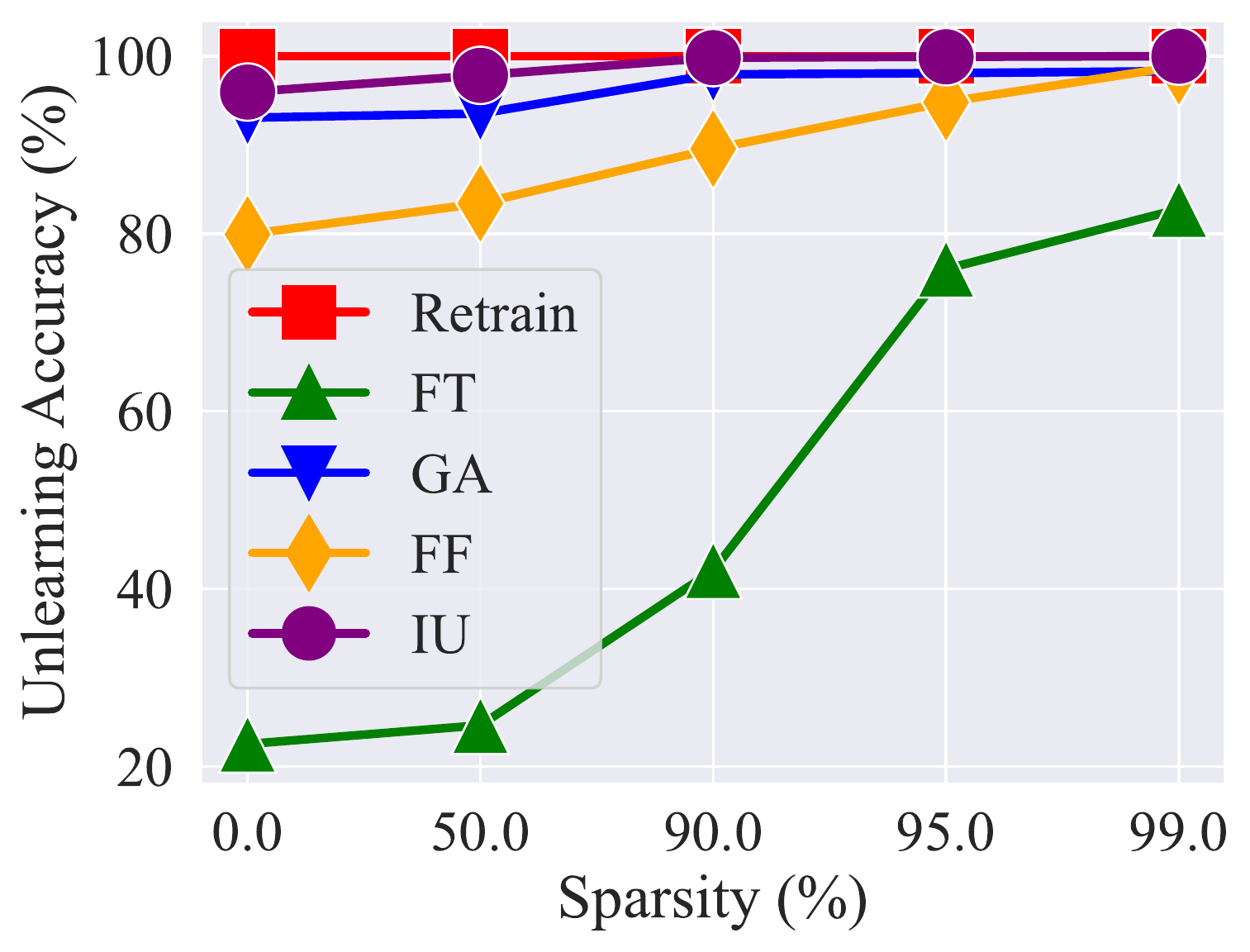} &
    \hspace*{-5mm} \includegraphics[width=0.25\textwidth,height=!]{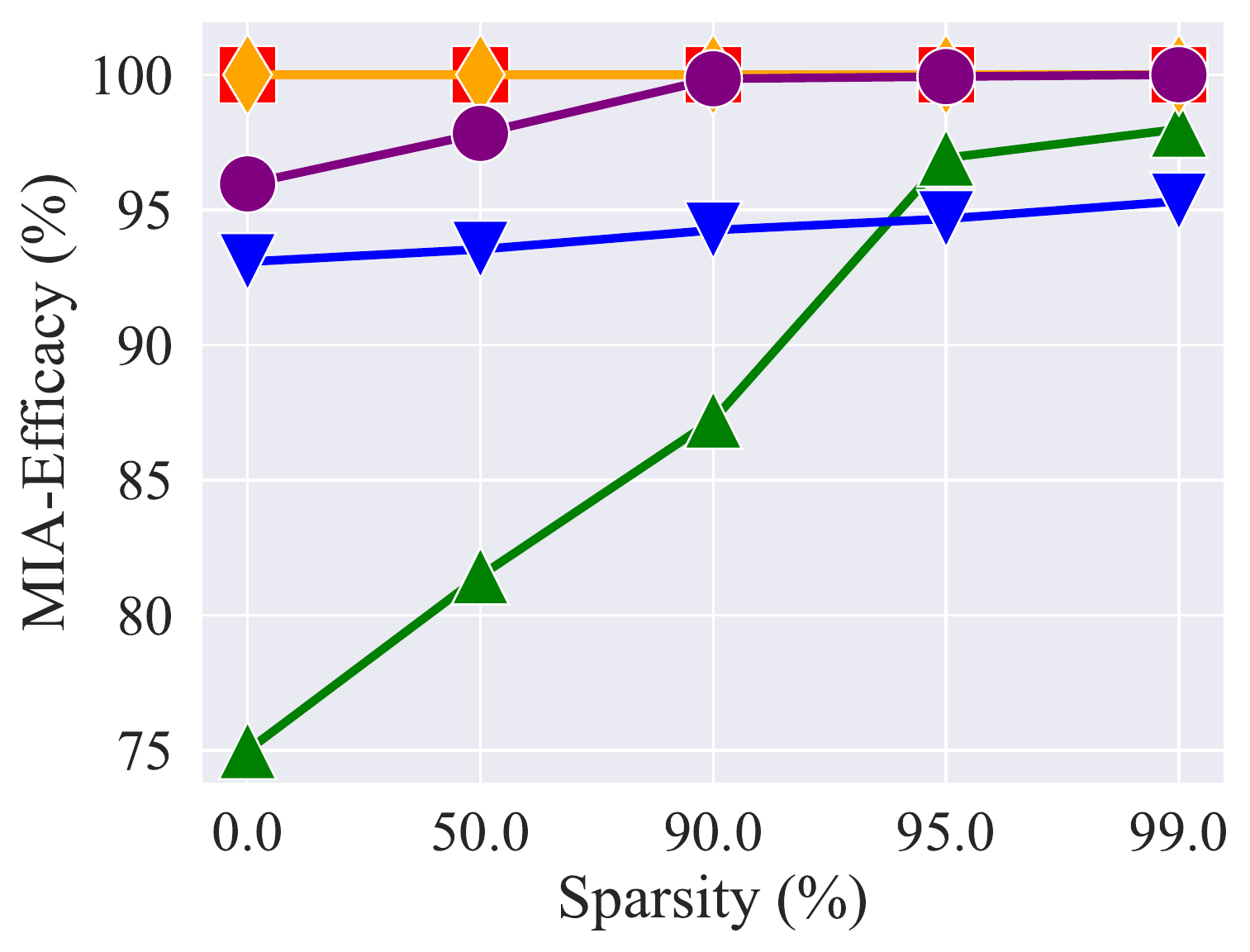} &
    \hspace*{-5mm} \includegraphics[width=0.25\textwidth,height=!]{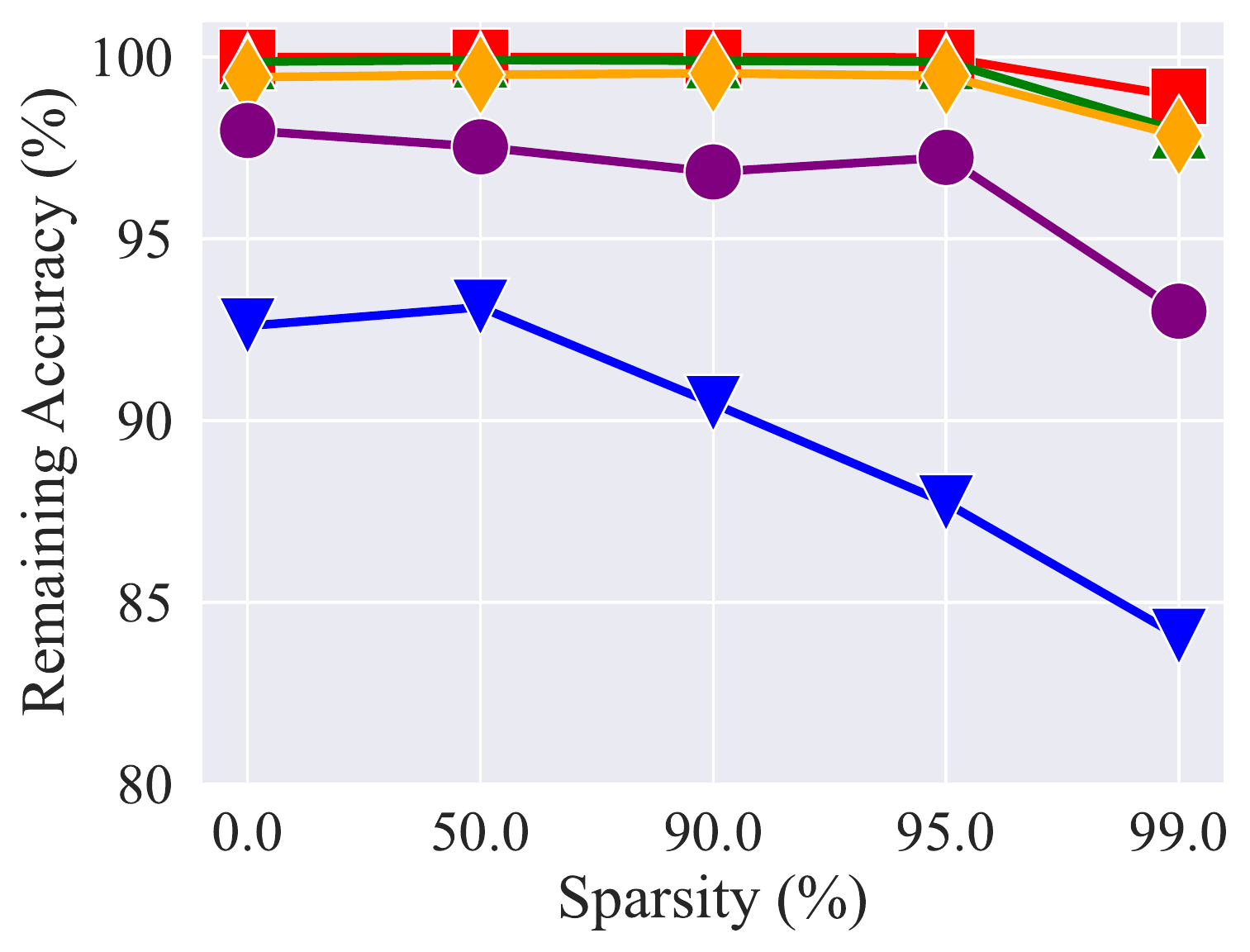} &
    \hspace*{-5mm}  \includegraphics[width=0.25\textwidth,height=!]{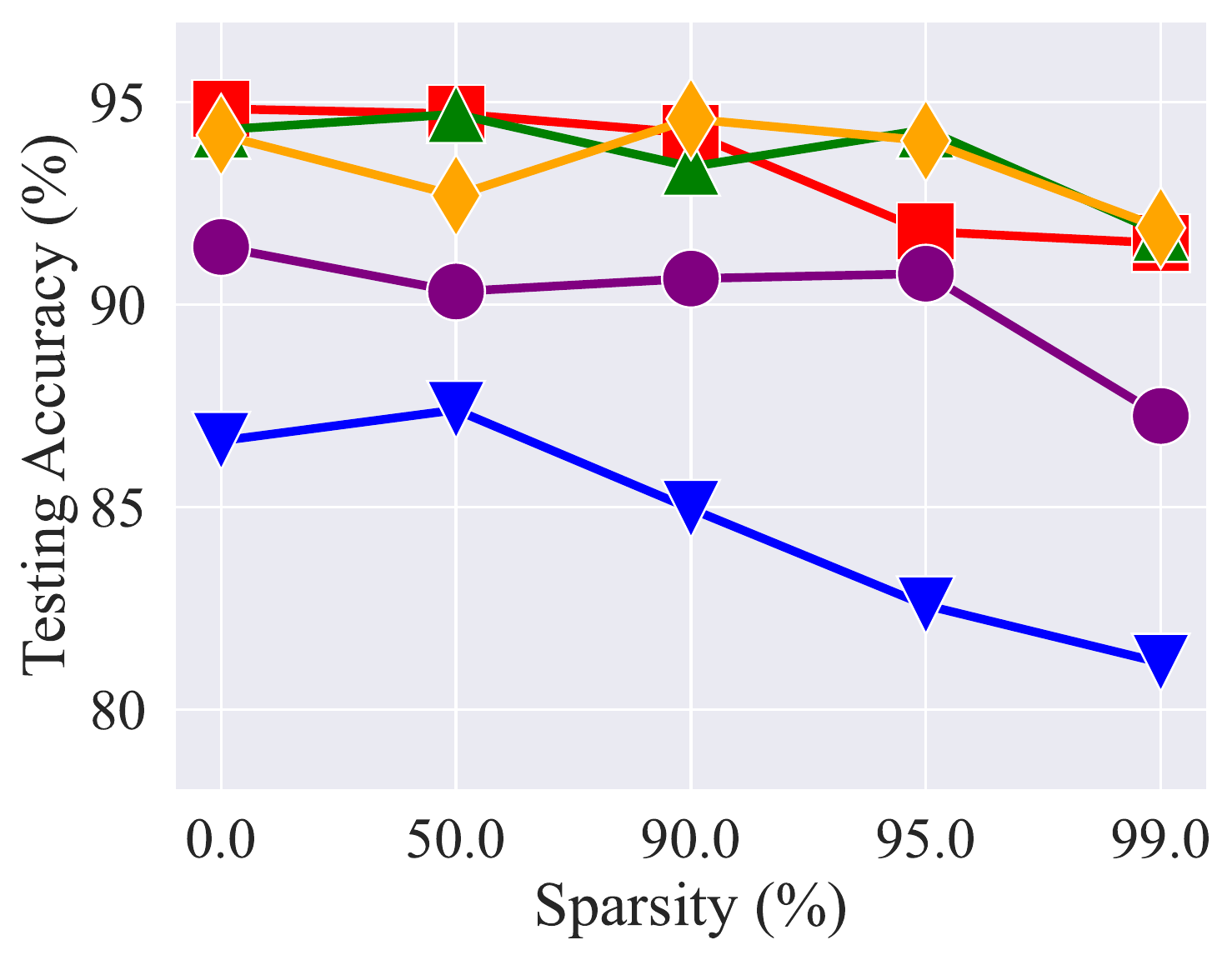} 
\end{tabular}}
 \vspace*{-3mm}
\caption{\footnotesize{
Performance of approximate unlearning ({\FTC}, {\GAC}, {\FFC}, {\IUC}) and exact unlearning ({\retrainC}) in efficacy ({\UA} and {\MIAF}), fidelity ({\RA}), and generalization ({\TA}) vs. model sparsity (achieved by {OMP}) in the data-model setup (CIFAR-10, ResNet-18). 
The unlearning scenario is class-wise 
 forgetting, and 
the average unlearning  performance over 10 classes is reported. We remark that being closer to \textcolor{red}{Retrain} performance is better for approximate MU schemes.
}}
\vspace*{-7mm}
\label{fig: results_OMP_MU}
\end{figure*}

Inspired by Prop.\,\ref{prop: SGD_sparse_MU}, we ask: 
\textit{Does the above   benefit of model sparsification in {\MU} apply to   other approximate unlearning methods besides {\GA}?} This drives us to investigate the performance of approximate unlearning across the entire spectrum as depicted in  {Tab.\,\ref{tab: summary_MU_methods_metrics}}.
Therefore,
\textbf{Fig.\,\ref{fig: results_OMP_MU}} shows the unlearning efficacy ({\UA} and {\MIAF}), fidelity ({\RA}), and generalization ({\TA}) of different   approximate unlearning methods   in the sparse model regime. Here class-wise forgetting is considered for {\MU} and OMP is used for weight pruning. 
As we can see, the efficacy of approximate unlearning is significantly improved as the model sparsity increases, \textit{e.g.},   {\UA} and {\MIAF} of using {\FT} over 90\% sparsity.  By contrast, {\FT} over the dense  model (0\% sparsity) is the least effective for {\MU}. 
Also, the efficacy gap between exact unlearning ({\retrain}) and  approximate unlearning  reduces on sparse models. Further, through the fidelity and generalization lenses, {\FT} and {\FF} yield the {\RA} and {\TA}  performance closest to {\retrain}, compared to other unlearning methods. In the regime of ultra-high  sparsity (99\%), 
the efficacy of unlearning exhibits a tradeoff with  {\RA} and {\TA} to some extent.   


    

\section{Sparsity-Aided Machine Unlearning}
\label{sec: sparsity_MU_alg}

Our study in Sec.\,\ref{sec: sparsityMU} suggests the new  {\MU} paradigm `prune first, then unlearn', which leverages the fact that  (approximate) unlearning on a sparse model yields a smaller unlearning error (Prop.\,\ref{prop: SGD_sparse_MU}) and improves the efficacy of {\MU}   (Fig.\,\ref{fig: results_OMP_MU}).
This promising finding, however, raises some new questions.  First, it remains elusive how the choice of a weight pruning method impacts the unlearning performance. Second,
it leaves room for developing  {sparsity-aware} {\MU} methods that can directly scrub data influence from a dense model.  

\noindent \textbf{Prune first, then unlearn: Choice of pruning methods.}
  There exist many ways to find the desired sparse model
in addition to OMP. Examples include pruning at random
initialization before training \cite{tanaka2020pruning,frankle2020pruning}
 and simultaneous pruning-training iterative magnitude pruning (\textbf{IMP}) \cite{frankle2018lottery}.
  Thus, the problem of pruning method selection arises for {\MU}. From the viewpoint of {\MU},  the unlearner would prioritize a pruning method that satisfies the following criteria:  \ding{182} \textit{least dependence}  on the forgetting dataset ($\Df$), \ding{183} \textit{lossless generalization} when pruning, and \ding{184} \textit{pruning efficiency}. The rationale behind \ding{182} is  that  
  it is desirable \textit{not} to incorporate    information of $\Df$ when seeking a sparse model prior to unlearning. 
  And the criteria \ding{183} and \ding{184} ensure that sparsity cannot hamper {\TA} (testing accuracy)  and {\RTE} (run-time efficiency). 
  Based on \ding{182}-\ding{184}, 
  we   propose to use two pruning methods.  

\noindent  \ding{70} \textbf{SynFlow} (synaptic flow pruning) \cite{tanaka2020pruning}: 
SynFlow provides a (training-free) pruning method at initialization, even without   accessing the dataset. 
Thus, it is uniquely suited for {\MU} to meet  the criterion \ding{182}. And SynFlow is easy to compute and yields a generalization improvement over many other pruning-at-initialization methods; see justifications in \cite{tanaka2020pruning}.

\noindent \ding{70} \textbf{OMP} (one-shot magnitude pruning)  \cite{ma2021sanity}: 
Different from SynFlow, {OMP}, which we focused on in Sec.\,\ref{sec: sparsityMU},
is performed over the original model   ($\thetafull$). It may depend on
the forgetting dataset ($\Df$), but has a much weaker dependence compared to {IMP}-based methods.
Moreover, {OMP} is  computationally lightest (\textit{i.e.}  best for \ding{184}) and    can yield better generalization than {SynFlow} \cite{zhang2022advancing}.

\begin{wrapfigure}{r}{72mm}
\vspace*{-4.5mm}
\centerline{
\begin{tabular}{cccc}
\hspace*{-5mm} \includegraphics[width=35mm,height=!]{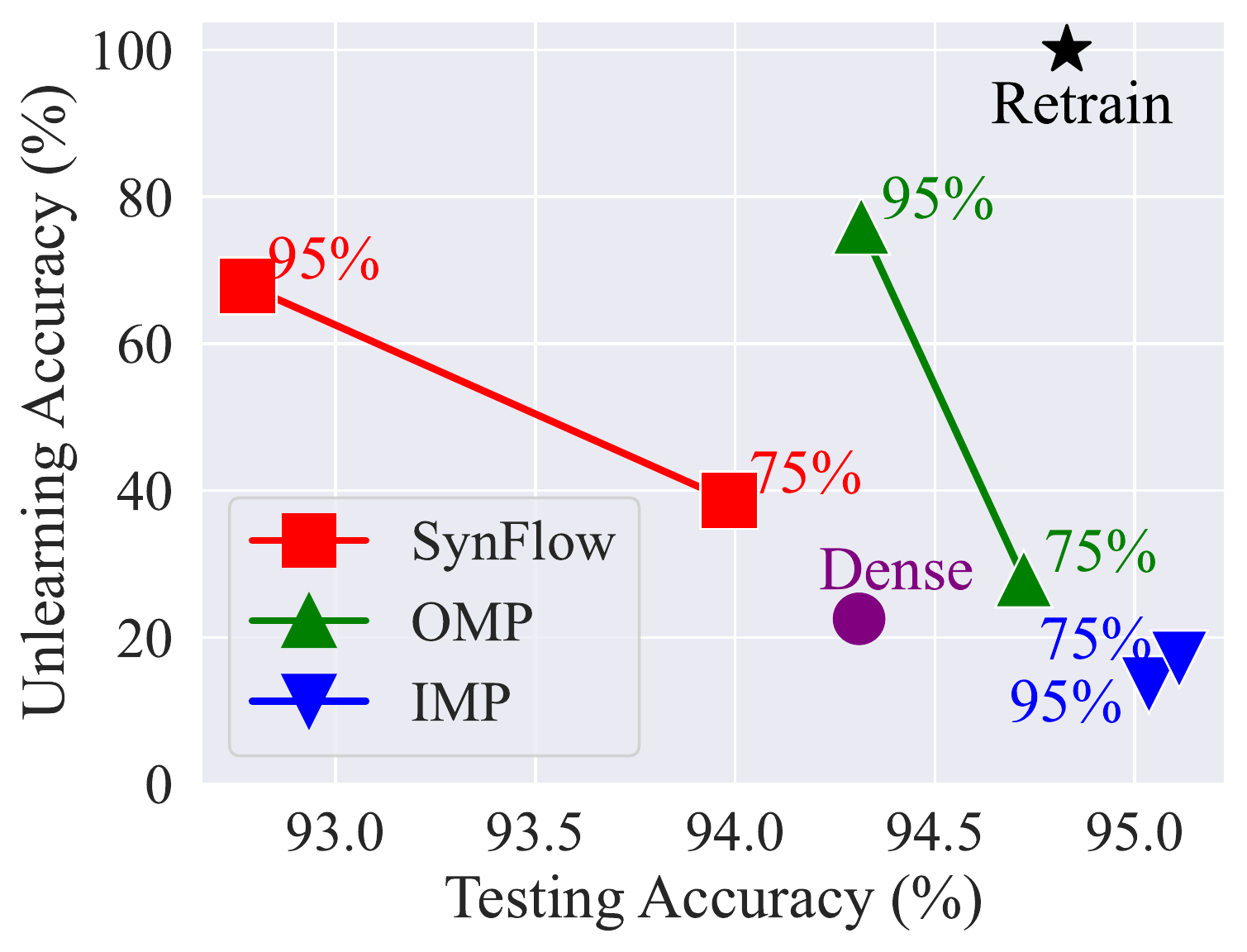} &
    \hspace*{-5mm}  \includegraphics[width=35mm,height=!]{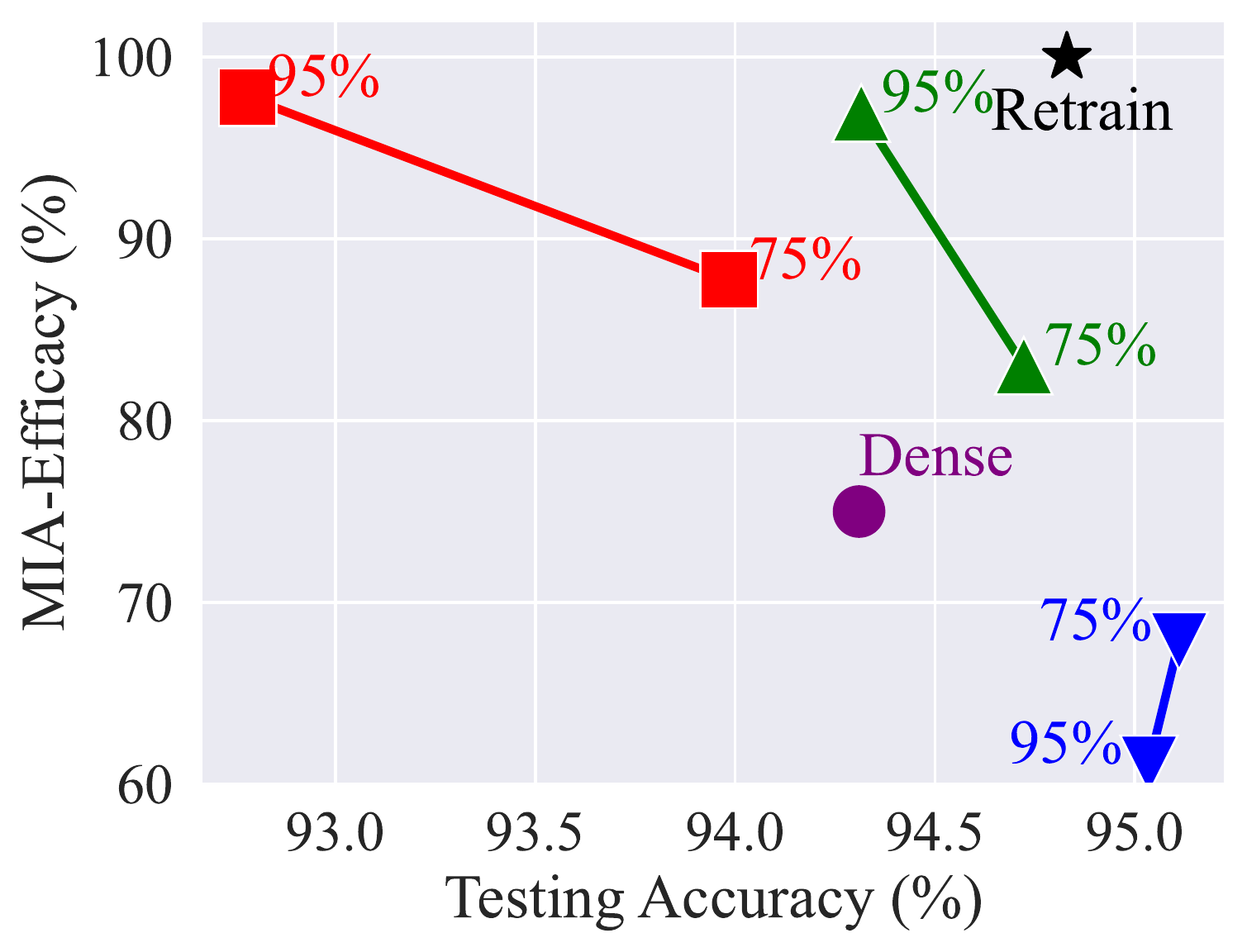} 
    \hspace*{-3mm} 
\end{tabular}}
 \vspace*{-3mm}
\caption{\footnotesize{Influence of different   pruning
methods ({\color{red}{SynFlow}}, {\color{ForestGreen}{OMP}}, and {\color{blue}{IMP}}) in unlearning efficacy ({\UA} and {\MIAF}) and generalization ({\TA})  on    (CIFAR-10, ResNet-18). \textbf{Left}: {\UA} vs. {\TA}. \textbf{Right}:   {\MIAF} vs. {\TA}. Each point is a {\FT}-based unlearned dense or sparse model (75\% or 95\% sparsity), or a retrained dense model.
}}
 \vspace*{-4mm}
\label{fig: results_pruning_comparison}
\end{wrapfigure}
Furthermore, it is important to clarify that IMP (iterative magnitude pruning) is \textit{not} suitable for {\MU}, despite being widely used to find the most accurate sparse models (\textit{i.e.}, best for criterion \ding{183}).
 Compared with the proposed pruning methods, IMP has the largest computation overhead   and the strongest correlation with the training dataset (including $\Df$), thereby deviating from \ding{182} and \ding{184}.
 In \textbf{Fig.\,\ref{fig: results_pruning_comparison}}, we show  the   efficacy  of {\FT}-based unlearning on sparse models generated  using  different pruning methods (SynFlow, OMP, and IMP). 
 As we can see, unlearning on {SynFlow} or {OMP}-generated sparse models yields improved {\UA} and {\MIAF} over that on the original dense model and   {IMP}-generated sparse models.  This unlearning improvement over the dense model is consistent with Fig.\,\ref{fig: results_OMP_MU}. More interestingly, we find that {IMP} \textit{cannot} benefit the unlearning efficacy, although it leads to the best {\TA}. This is because {IMP} heavily relies on the training set including forgetting data points, which is revealed by the empirical results -- the unlearning metrics get worse for IMP with increasing sparsity.
 Furthermore, when examining the performance of SynFlow and OMP, we observe that the latter   generally outperforms the former, exhibiting results that are closer to those of {\retrain}. 
 Thus,  \textit{{OMP} is the pruning method we will use by default}.

 
 





{\noindent \textbf{Sparsity-aware  unlearning.}}
We next study if pruning and unlearning can be carried out simultaneously, without requiring prior knowledge of model sparsity. Let $\Lunl (\btheta; \thetafull,  \Dr)$ denote the unlearning objective function of model parameters $\btheta$, given the pre-trained  state $\thetafull$, 
and the remaining training dataset $\Dr$. 
Inspired by  sparsity-inducing optimization \cite{bach2012optimization}, we integrate an $\ell_1$ norm-based sparse penalty into  $\Lunl $. This leads to the problem of `\textbf{\MUSparse}':

\vspace*{-3mm}
{\small{\begin{align}
    \thetaunl = \argmin_{\btheta} \Lunl (\btheta; \thetafull,  \Dr) + \gamma \| \btheta \|_1,
    \label{eq: MUSparse}
\end{align}}}%
where we specify $\Lunl$ by the fine-tuning objective, and $\gamma > 0$ is a regularization parameter that {controls the penalty level of the $\ell_1$ norm, thereby reducing the magnitudes of `unimportant' weights.}

\vspace{-5mm}
\begin{table}[htb!]
\centering
\caption{\footnotesize{{\MU} performance  comparison of using {\MUSparse} with different   sparsity schedulers of $\gamma$  in \eqref{eq: MUSparse} and using {\retrain}.  
The unlearning scenario is given by random data forgetting (10\% data points across all classes) on (ResNet-18, CIFAR-10).
A performance gap  against \textcolor{blue}{{\retrain}} is provided 
in (\textcolor{blue}{$\bullet$}).
}}
\label{tab: ablation_l1_scheduler}
\vspace*{1mm}
\resizebox{0.85\textwidth}{!}{
\begin{tabular}{c|c|c|c|c|c}
\toprule[1pt]
\midrule
  {\MU}& {\UA} & {{\MIAF}}& {{\RA}} & {{\TA}} & {{RTE} (min)} \\ 


\midrule
{\retrain} & \textcolor{blue}{5.41} & \textcolor{blue}{13.12} & \textcolor{blue}{100.00} & \textcolor{blue}{94.42} & 42.15
\\
{\MUSparse} + constant $\gamma$ & 6.60 (\textcolor{blue}{1.19}) & 14.64 (\textcolor{blue}{1.52}) & 96.51 (\textcolor{blue}{3.49})	& 87.30 (\textcolor{blue}{7.12}) & 2.53
\\

{\MUSparse} + linear growing $\gamma$  & 3.80 (\textcolor{blue}{1.61}) & 8.75 (\textcolor{blue}{4.37}) & 97.13 (\textcolor{blue}{2.87})	& 90.63 (\textcolor{blue}{3.79}) & 2.53
\\
{\MUSparse} + linear decaying $\gamma$ & \textbf{5.35} (\textcolor{blue}{\textbf{0.06}}) & \textbf{12.71} (\textcolor{blue}{\textbf{0.41}}) & \textbf{97.39} (\textcolor{blue}{\textbf{2.61}})	& {\textbf{91.26}} (\textcolor{blue}{\textbf{3.16}}) & 2.53
\\
\midrule
\bottomrule[1pt]
\end{tabular}
}
\end{table}
In practice,  the unlearning performance could be sensitive to the choice of the sparse regularization parameter $\gamma$. To address this limitation, we propose the design of a sparse regularization scheduler. Specifically, we explore three schemes: (1) constant $\gamma$, (2) linearly growing $\gamma$ and (3) linearly decaying $\gamma$; see Sec.\,\ref{sec: exp_setup} for detailed implementations. Our empirical evaluation presented in \textbf{Tab.\,\ref{tab: ablation_l1_scheduler}} shows that the use of a linearly decreasing $\gamma$ scheduler outperforms   other schemes.  
This scheduler not only minimizes the gap in unlearning efficacy compared to {\retrain}, but also improves the preservation of {\RA} and {\TA} after unlearning. 
These findings suggest that it is advantageous to prioritize promoting sparsity during the early stages of unlearning and then gradually shift the focus towards enhancing fine-tuning accuracy on the remaining dataset $\Dr$.

\section{Experiments}
\label{sec: exp}



\subsection{Experiment setups}
\label{sec: exp_setup}

\noindent \textbf{Datasets and models.}
Unless specified otherwise, our experiments will focus  on image classification under 
 CIFAR-10 \cite{krizhevsky2009learning}  using ResNet-18 \cite{he2016deep}. Yet,  
 experiments  on 
  additional datasets (CIFAR-100 \cite{krizhevsky2009learning}, SVHN \cite{netzer2011reading}, and ImageNet \cite{deng2009imagenet}) and an  alternative model architecture (VGG-16 \cite{simonyan2014very}) can  be found in Appendix\,\ref{appendix: additional results}.
  Across all the aforementioned datasets and model architectures, our experiments consistently show that model sparsification can effectively reduce the gap between approximate unlearning and exact unlearning. 
 
 



\noindent \textbf{Unlearning and pruning setups.}
 We   focus on two unlearning scenarios mentioned in Sec.\,\ref{sec: primer_MU}, \textit{class-wise forgetting} and \textit{random data forgetting} ({$10\%$ of the whole training dataset} {together with 10 random trials}). 
In the `\textit{prune first, then unlearn}' paradigm, we   focus on unlearning methods ({\FT}, {\GA}, {\FF},  and {\IU}) shown in Tab.\,\ref{tab: summary_MU_methods_metrics}
when applying to sparse models.
We  implement these methods following their official repositories. However, it is worth noting that  the  implementation of {\FF} in \citet{golatkar2020eternal}
modifies the model architecture in class-wise forgetting, \textit{i.e.}, removes the  prediction head  of the class to be scrubbed. 
By contrast, other   methods   keep the model architecture  intact during unlearning. 
Also, we choose {OMP} as the  default pruning method, 
as justified in Fig.\,\ref{fig: results_pruning_comparison}.
In the `\textit{sparsity-aware unlearning}' paradigm, the sparsity-promoting regularization parameter $\gamma$ in \eqref{eq: MUSparse} is determined through the line search in the interval $[10^{-5}, 10^{-1}]$, with consideration for the trade-off between testing accuracy and unlearning accuracy. For all schedulers, $\gamma$ is set around to $5 \times 10^{-4}$. The linearly increasing and decaying schedulers are implemented as $\gamma_t = \frac{2t}{T} \gamma$ and $\gamma_t = (2 - \frac{2t}{T})\gamma$ respectively, where $t$ is the epoch index and $T$ is the total number of epochs. 
We refer readers to Appendix\,\ref{appendix: training and unlearning settings} for more details. 

\noindent \textbf{Evaluation metrics.}
We evaluate the unlearning performance following Tab.\,\ref{tab: summary_MU_methods_metrics}. 
Recall that {\UA}   and {\MIAF}   depict the \textit{efficacy} of {\MU}, {\RA}  reflects the \textit{fidelity} of {\MU}, and {\TA}   and {\RTE}  characterize the \textit{generalization ability} and the \textit{computation efficiency} of  an unlearning method. 
We implement MIA (membership inference attack) using the prediction confidence-based attack method \cite{song2019privacy,yeom2018privacy}, whose effectiveness has been justified in \citet{song2020systematic} compared to other   methods. We refer readers to Appendix\,\ref{appendix: metric settings} for more implementation details. {To more precisely gauge the proximity of each approximate {\MU} to {\retrain}, we introduce a metric termed `Disparity Average'. This metric quantifies the mean performance gap between each unlearning method and {\retrain} across all considered metrics. A lower value indicates closer performance to {\retrain}.}



\subsection{Experiment results}
\label{sec: experiment_results}
\begin{table*}[htb!]
\centering
\caption{ Performance overview of various MU methods  on dense and 95\%-sparse models considering different unlearning scenarios:
 class-wise forgetting, 
 and random data forgetting. The forgetting data of random data forgetting ratio is $10\%$ of the whole training dataset, 
 the sparse models are obtained using OMP \cite{ma2021sanity}, and the unlearning methods and evaluation metrics are summarized in Tab.\,\ref{tab: summary_MU_methods_metrics}. {Class-wise forgetting is conducted class-wise.}
The performance is reported in the form $a_{\pm b}$, with mean $a$ and standard deviation $b$ computed over $10$ independent trials. 
A performance gap  against \textcolor{blue}{{\retrain}} is provided 
in (\textcolor{blue}{$\bullet$}). Note that the better performance of approximate unlearning corresponds to the smaller performance gap with the gold-standard retrained model. {`Disparity Ave.' represents the average unlearning gaps across diverse metrics.}
}
\vspace*{-2mm}
\label{tab: overall_performance}
\resizebox{0.98\textwidth}{!}{
\begin{tabular}{c|cc|cc|cc|cc|cc|c}
\toprule[1pt]
\midrule
  \multirow{2}{*}{\MU}& \multicolumn{2}{c|}{{\UA}} & \multicolumn{2}{c|}{{\MIAF}}& \multicolumn{2}{c|}{{\RA}} & \multicolumn{2}{c|}{{\TA}}&\multicolumn{2}{c|}{{Disparity Ave. $\downarrow$}}& {\RTE}  \\ 
  & \multicolumn{1}{c|}{{\textsc{Dense}}}  & \multicolumn{1}{c|}{$\mathbf{95\%}$ \textbf{Sparsity}}
    & \multicolumn{1}{c|}{\textsc{Dense}}  & \multicolumn{1}{c|}{$\mathbf{95\%}$ \textbf{Sparsity}}
    & \multicolumn{1}{c|}{\textsc{Dense}}  & \multicolumn{1}{c|}{$\mathbf{95\%}$ \textbf{Sparsity}}
      & \multicolumn{1}{c|}{\textsc{Dense}}  & \multicolumn{1}{c|}{$\mathbf{95\%}$ \textbf{Sparsity}} & \multicolumn{1}{c|}{\textsc{Dense}}  & \multicolumn{1}{c|}{$\mathbf{95\%}$ \textbf{Sparsity}} & (min)
  \\

\midrule
\rowcolor{Gray}
\multicolumn{12}{c}{Class-wise forgetting} \\
\midrule
\retrain &\textcolor{blue}{$100.00_{\pm{0.00}}$}    & \textcolor{blue}{$100.00_{\pm{0.00}}$}
&\textcolor{blue}{$100.00_{\pm{0.00}}$}   & \textcolor{blue}{$100.00_{\pm{0.00}}$}
&\textcolor{blue}{$100.00_{\pm{0.00}}$}    & \textcolor{blue}{$99.99_{\pm{0.01}}$}
&\textcolor{blue}{$94.83_{\pm{0.11}}$}   & \textcolor{blue}{$91.80_{\pm{0.89}}$} & 0.00 & 0.00
 &43.23\\
  \FT &$22.53_{\pm{8.16}}$ (\textcolor{blue}{$77.47$})&${73.64}_{\pm{9.46}}$  (\textcolor{blue}{${26.36}$})&$75.00_{\pm{14.68}}$ (\textcolor{blue}{${25.00}$})& ${83.02}_{\pm{16.33}}$ (\textcolor{blue}{${16.98}$}) 
  &$99.87_{\pm{0.04}}$ (\textcolor{blue}{$0.13$}) & ${99.87}_{\pm{0.05}}$ (\textcolor{blue}{${0.12}$})&$94.31_{\pm{0.19}}$ (\textcolor{blue}{$0.52$})
 &$94.32_{\pm{0.12}}$ (\textcolor{blue}{$2.52$})
&  25.78&11.50& 2.52

  \\
 \GA &$93.08_{\pm{2.29}}$ (\textcolor{blue}{6.92}) &${98.09}_{\pm{1.11}}$ (\textcolor{blue}{${1.91}$})
& $94.03_{\pm{3.27}}$ (\textcolor{blue}{5.97})& ${97.74}_{\pm{2.24}}$ (\textcolor{blue}{${2.26}$})
& $92.60_{\pm{0.25}}$ (\textcolor{blue}{$7.40$})& $87.74_{\pm{0.27}}$ (\textcolor{blue}{$12.25$}) 
& $86.64_{\pm{0.28}}$ (\textcolor{blue}{$8.19$})& $82.58_{\pm{0.27}}$ (\textcolor{blue}{$9.22$}) 
&7.12 &6.41&  0.33
 \\
  {\FF}  & $79.93_{\pm{8.92}}$ (\textcolor{blue}{$20.07$})& ${94.83}_{\pm{4.29}}$ (\textcolor{blue}{${5.17}$}) 
  & $100.00_{\pm{0.00}}$ (\textcolor{blue}{$0.00$})& ${100.00}_{\pm{0.00}}$ (\textcolor{blue}{$0.00$}) 
    & $99.45_{\pm{0.24}}$ (\textcolor{blue}{$0.55$})& ${99.48}_{\pm{0.33}}$ (\textcolor{blue}{${0.51}$})
        & $94.18_{\pm{0.08}}$ (\textcolor{blue}{$0.65$})& $94.04_{\pm{0.10}}$ (\textcolor{blue}{$2.24$})&5.32 &1.98&38.91
  \\
 \IU 
  &$87.82_{\pm{2.15}} $ (\textcolor{blue}{$12.18$})& ${99.47}_{\pm{0.15}}$ (\textcolor{blue}{${0.53}$})
 & $95.96_{\pm0.21}$ (\textcolor{blue}{$4.04$})
&${99.93}_{\pm{0.04}}$ (\textcolor{blue}{${0.07}$})
 &$97.98_{\pm{0.21}}$ (\textcolor{blue}{$2.02$}) 
 &$97.24_{\pm{0.13}}$ (\textcolor{blue}{$2.75$}) 
 &$91.42_{\pm{0.21}}$ (\textcolor{blue}{$3.41$})&${90.76_{\pm{0.18}}}$ (\textcolor{blue}{${1.04}$}) &5.41&1.10& 3.25
 \\

\midrule
\rowcolor{Gray}
\multicolumn{12}{c}{Random data forgetting} \\
\midrule
 \retrain &\textcolor{blue}{$5.41_{\pm{0.11}}$}&\textcolor{blue}{$ 6.77_{\pm{0.23}}$}&\textcolor{blue}{$13.12_{\pm{0.14}}$}&\textcolor{blue}{$14.17_{\pm{0.18}}$}&\textcolor{blue}{$100.00_{\pm{0.00}}$}&\textcolor{blue}{$100.00_{\pm{0.00}}$}&\textcolor{blue}{$94.42_{\pm{0.09}}$}&\textcolor{blue}{$93.33_{\pm{0.12}}$} & 0.00 & 0.00 & 42.15 
\\
 \FT & $6.83_{\pm{0.51}}$ (\textcolor{blue}{$1.42$})& $5.97_{\pm{0.57}}$ (\textcolor{blue}{$0.80$})& $14.97_{\pm{0.62}}$ (\textcolor{blue}{$1.85$})& $13.36_{\pm{0.59}}$ (\textcolor{blue}{$0.81$})& $96.61_{\pm{0.25}}$ (\textcolor{blue}{$3.39$})& $96.99_{\pm{0.31}}$ (\textcolor{blue}{$3.01$})& $90.13_{\pm{0.26}}$ (\textcolor{blue}{$4.29$})& $90.29_{\pm{0.31}}$ (\textcolor{blue}{$3.04$}) & 2.74 & 1.92 & 2.33  
 \\
 \GA & $7.54_{\pm{0.29}}$ (\textcolor{blue}{$2.13$})& $5.62_{\pm{0.46}}$ (\textcolor{blue}{$1.15$})& $10.04_{\pm{0.31}}$ (\textcolor{blue}{$3.08$})& $11.76_{\pm{0.52}}$ (\textcolor{blue}{$2.41$})& $93.31_{\pm{0.04}}$ (\textcolor{blue}{$6.69$})& $95.44_{\pm{0.11}}$ (\textcolor{blue}{$4.56$})& $89.28_{\pm{0.07}}$ (\textcolor{blue}{$5.14$})& $89.26_{\pm{0.15}}$ (\textcolor{blue}{$4.07$}) & 4.26 & 3.05 & 0.31
 \\
  \FF & $7.84_{\pm{0.71}}$ (\textcolor{blue}{$2.43$})& $8.16_{\pm{0.67}}$ (\textcolor{blue}{$1.39$})& $9.52_{\pm{0.43}}$ (\textcolor{blue}{$3.60$})& $10.80_{\pm{0.37}}$ (\textcolor{blue}{$3.37$})& $92.05_{\pm{0.16}}$ (\textcolor{blue}{$7.95$})& $92.29_{\pm{0.24}}$ (\textcolor{blue}{$7.71$})& $88.10_{\pm{0.19}}$ (\textcolor{blue}{$6.32$})& $87.79_{\pm{0.23}}$ (\textcolor{blue}{$5.54$}) &5.08 & 4.50 & 38.24
 \\
  \IU & $2.03_{\pm{0.43}}$ (\textcolor{blue}{$3.38$})& $6.51_{\pm{0.52}}$ (\textcolor{blue}{$0.26$})& $5.07_{\pm{0.74}}$ (\textcolor{blue}{$8.05$})& $11.93_{\pm{0.68}}$ (\textcolor{blue}{$2.24$})& $98.26_{\pm{0.29}}$ (\textcolor{blue}{$1.74$})& $94.94_{\pm{0.31}}$ (\textcolor{blue}{$5.06$})& $91.33_{\pm{0.22}}$ (\textcolor{blue}{$3.09$})& $88.74_{\pm{0.42}}$ (\textcolor{blue}{$4.59$}) &4.07 & 3.08& 3.22 \\
\midrule
\bottomrule[1pt]
\end{tabular}
}
\vspace*{-4mm}
\end{table*}

\noindent \textbf{Model sparsity improves approximate unlearning.}
In \textbf{Tab.\,\ref{tab: overall_performance}}, we study the impact of model sparsity  on  the performance of various {\MU} methods 
in the `prune first, then unlearn' paradigm. 
The performance of the exact unlearning method  ({\retrain}) is also provided for comparison. 
Note that the better performance of    approximate unlearning  corresponds to the  smaller performance gap with  the gold-standard retrained model. 

\textit{First}, given an approximate unlearning  method ({\FT}, {\GA}, {\FF}, or {\IU}), we consistently observe that model sparsity improves {\UA} and {\MIAF} (\textit{i.e.}, the efficacy of approximate unlearning) without  much performance loss in  {\RA} (\textit{i.e.}, fidelity).
In particular, 
  the  performance gap between each approximate unlearning method and {\retrain} reduces as the model becomes sparser (see the `95\% sparsity' column vs. the `dense' column).
  Note that 
  the performance gap against {\retrain} is highlighted in $(\cdot)$ for each approximate unlearning.
  We also   observe that {\retrain}  on the 95\%-sparsity model encounters a  3\% {\TA} drop. Yet, from the perspective of approximate unlearning, this drop brings in 
  a more significant improvement  in    {\UA} and {\MIAF}  when model sparsity is promoted. Let us take {\FT} (the simplest unlearning method) for class-wise forgetting as an example. As the model sparsity   reaches   $95\%$, we obtain $51\%$ {\UA} improvement and $8\%$ {\MIAF} improvement. 
  Furthermore, {\FT} and {\IU} on the 95\%-sparsity model can better preserve  {\TA} compared to other  methods. 
Table\,\ref{tab: overall_performance} further indicates that sparsity reduces average disparity compared to a dense model across various approximate {\MU} methods and unlearning scenarios.

\begin{wrapfigure}{r}{60mm}
\vspace*{-7.8mm}
\begin{center}
\includegraphics[width=60mm,height=!]{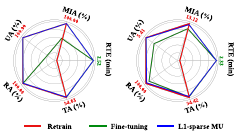}

\vspace*{-1mm}
\begin{tabular}{cc}
{\scriptsize{\hspace*{-1.5mm}(a) Class-wise forgetting}} &{\scriptsize{\hspace*{3mm}(b) Random data forgetting}}
\end{tabular}
\end{center}
\vspace*{-2.5mm}

\caption{\footnotesize{
 Performance   of sparsity-aware unlearning vs. {\FT} and {\retrain} on class-wise forgetting and random data forgetting under (CIFAR-10, ResNet-18). 
 Each   metric is normalized to $[0,1]$ based on the best result  across unlearning methods 
  for ease of visualization, while the actual best value  is provided (\textit{e.g.}, $2.52$  is the  least computation time for class-wise forgetting). 
}
}
  \label{fig: results_l1_sparse_unlearn}
\vspace*{-7mm}
\end{wrapfigure}

\textit{Second},  existing approximate unlearning methods have different pros and cons. Let us focus on the regime of $95\%$ sparsity.  We observe that {\FT}  typically yields the best {\RA} and {\TA}, which has a tradeoff with its unlearning efficacy ({\UA} and {\MIAF}). Moreover, {\GA} yields the worst {\RA} since it is most loosely connected with the remaining dataset $\Dr$.  {{\FF} becomes ineffective when scrubbing random data points compared to its class-wise unlearning performance. 
Furthermore,    {\IU} causes a {\TA} drop but yields the smallest gap with exact unlearning across diverse metrics under the $95\%$ model sparsity.
In Appendix\,\ref{appendix: additional results}, we provide additional results on CIFAR-100 and SVHN datasets, as shown in Tab.\,\ref{tab: overall_performance_ext_datasets}, as well as on the ImageNet dataset, depicted in Tab.\,\ref{tab: overall_performance_ImageNet}. Other results pertaining to the VGG-16 architecture are  provided in Tab.\,\ref{tab: overall_performance_ext_archs}.

\noindent \textbf{{Effectiveness of sparsity-aware unlearning.}}
In \textbf{Fig.\,\ref{fig: results_l1_sparse_unlearn}},
we   showcase the effectiveness of the proposed sparsity-aware unlearning method, \textit{i.e.}, {\MUSparse}. 
For ease of presentation, we focus on the comparison with  {\FT} and the optimal {\retrain}  strategy in both class-wise forgetting and random data forgetting  scenarios under (CIFAR-10, ResNet-18). As we can see, {\MUSparse}  outperforms {\FT} in  the unlearning efficacy ({\UA} and {\MIAF}), and closes the performance gap with {\retrain}  without losing the computation advantage of approximate unlearning. We refer readers to Appendix\,\ref{appendix: additional results} and Fig.\,\ref{fig: results_l1_sparse_unlearn_others} for further exploration of {\MUSparse} on additional datasets.

\begin{wrapfigure}{r}{80mm}
\vspace*{-3mm}
\centerline{
\begin{tabular}{cccc}
    \hspace*{-2mm}  \includegraphics[width=40mm,height=!]{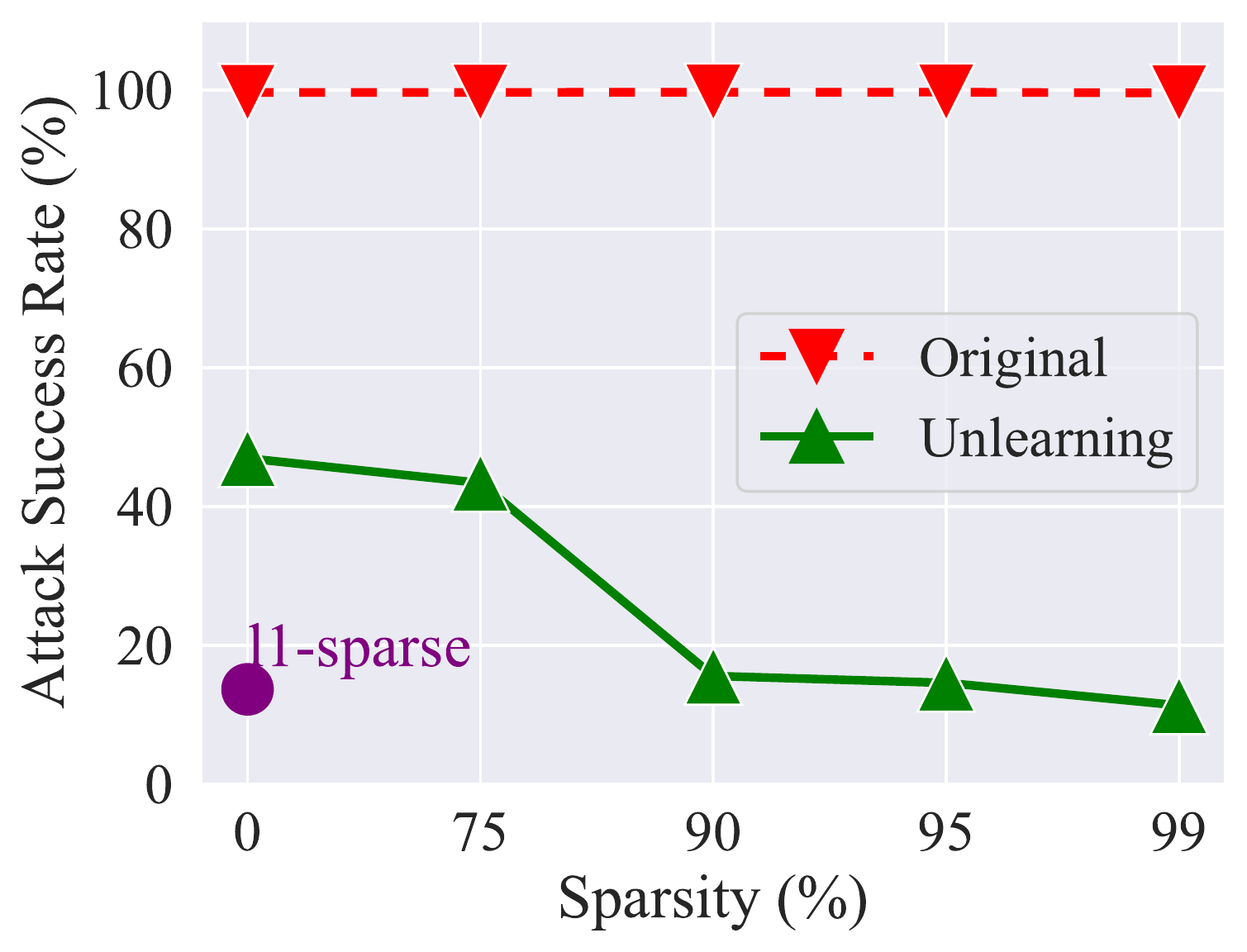} &
    \hspace*{-5mm} \includegraphics[width=40mm,height=!]{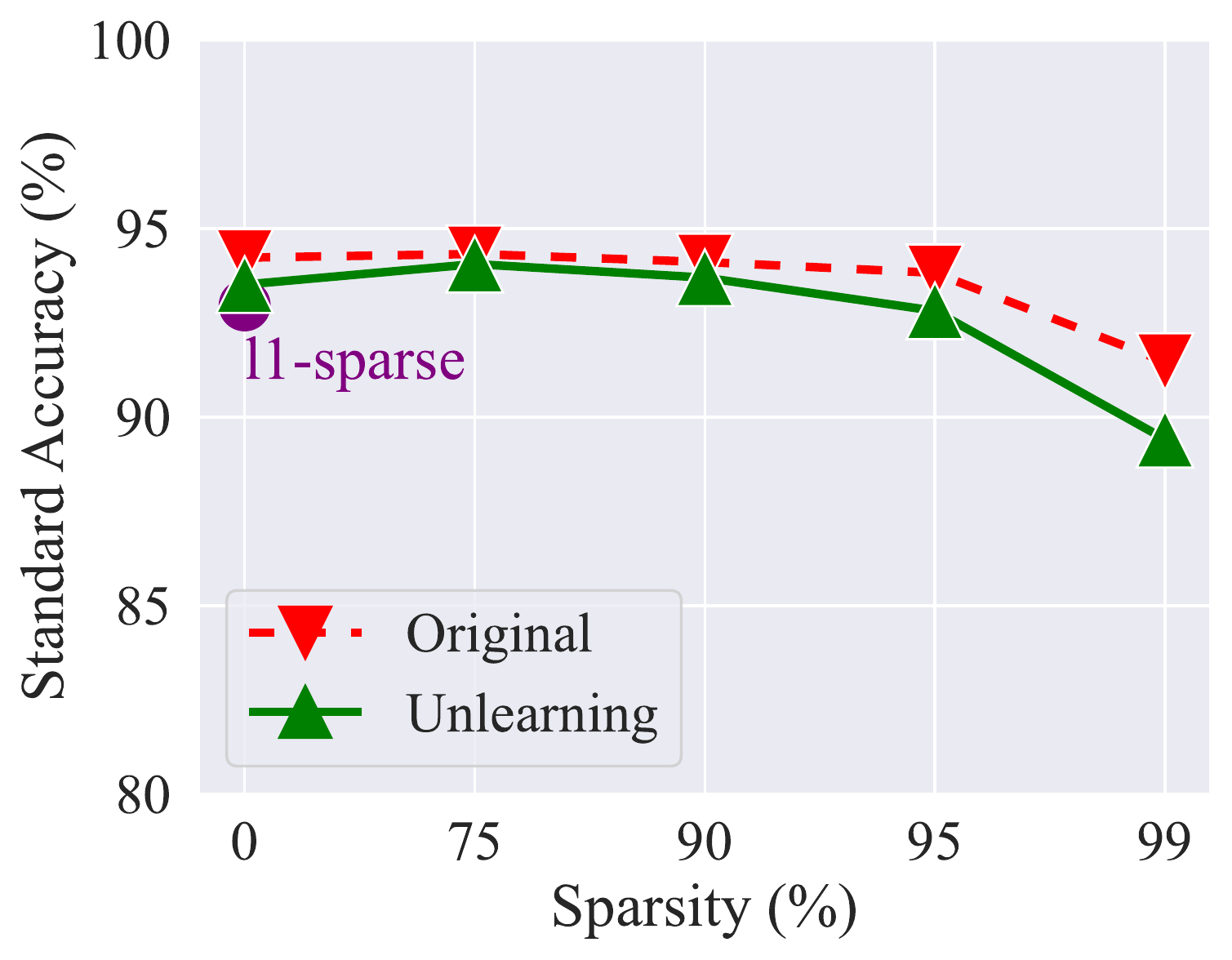} \\

\end{tabular}
}
\vspace*{-2mm}
\caption{
Performance  of  Trojan model cleanse   via proposed unlearning vs. model sparsity, where `Original' refers to the original Trojan model.
\textbf{Left}: ASR vs. model sparsity. \textbf{Right}: SA vs. model sparsity. 
}
  \label{fig: results_MU_pruning_backdoor}
 \vspace*{-4mm}
\end{wrapfigure}
\noindent \textbf{Application: {\MU} for Trojan model cleanse.}
We next present an application of {\MU} to remove the influence of poisoned backdoor data from a learned model,  following the backdoor attack setup   \cite{gu2017badnets}, where an adversary 
manipulates a small portion of training data (\textit{a.k.a.}   poisoning ratio) by 
injecting a backdoor trigger (\textit{e.g.}, a small image patch) and modifying data labels towards a targeted incorrect label.  
The trained model is called \textit{Trojan model}, yielding the backdoor-designated incorrect prediction if the trigger is present at testing. Otherwise, it behaves normally. 

We then regard {\MU} as a defensive method to scrub the harmful influence of  poisoned training data in  the model's prediction, with a similar motivation as \citet{liu2022backdoor}.
We evaluate the performance of the unlearned model from two perspectives, backdoor attack success rate (\textbf{ASR}) and standard accuracy (\textbf{SA}). 
\textbf{Fig.\,\ref{fig: results_MU_pruning_backdoor}} shows   ASR and {SA} of the   Trojan model (with poisoning ratio $10\%$)  and its unlearned version using the simplest {\FT} method against model sparsity. {Fig.\,\ref{fig: results_MU_pruning_backdoor} also includes the $\ell_1$-sparse {\MU} to demonstrate its effectiveness on  model cleanse. Since it is applied to a dense model (without using hard thresholding to force weight sparsity), it contributes just a single data point at the sparsity level 0\%.}
As we can see, the  original Trojan model maintains $100\%$ ASR and a similar SA across different model sparsity levels. By contrast, {\FT}-based unlearning can  reduce ASR without inducing much {SA} loss. Such a defensive advantage becomes more significant when sparsity reaches $90\%$. {Besides, $\ell_1$-sparse {\MU} can also effectively remove the backdoor effect while largely preserving the model’s generalization.} 
Thus, our proposed unlearning shows promise in  application of backdoor attack defense.

\noindent \textbf{Application: {\MU} to improve transfer learning.}
Further, we utilize the   {\MUSparse} method to mitigate the   impact of harmful data classes of   ImageNet    on transfer learning.  This approach is inspired by \citet{jain2022data}, which shows that  removing specific negatively-influenced ImageNet classes and retraining a source model  can enhance its transfer learning accuracy on    downstream   datasets  after finetuning. However, retraining the source model introduces additional computational overhead. {\MU} naturally addresses this limitation and offers a solution.


\textbf{Tab.\,\ref{tab: transfer_results}} illustrates the transfer learning accuracy of the unlearned or retrained source model (ResNet-18) on ImageNet, with $n$ classes removed. The downstream target datasets used for evaluation are  SUN397 \cite{xiao2010sun} and OxfordPets \cite{parkhi2012cats}.
The  employed finetuning approach   is linear probing, which finetunes the classification head of the source model on target datasets while keeping the feature extraction network of the source model intact. 
As we can see, removing data classes from the source ImageNet dataset    can lead to improved transfer learning accuracy compared to the conventional method of using the pre-trained model on the full ImageNet  (\textit{i.e.}, $n = 0$). Moreover,
our proposed 
\begin{wraptable}{r}{63mm}
\centering
\vspace*{-3.3mm}
\caption{
Transfer learning accuracy (Acc) and computation time (mins) of the unlearned   ImageNet model with $n \in \{ 100,200,300\}$ classes removed, where SUN397 and OxfordPets are downstream target datasets on linear probing transfer learning setting. When $n = 0$, transfer learning is performed using the pretrained model on the full ImageNet, serving as a baseline, together with the method in \cite{jain2022data}  for comparison. 
}
\label{tab: transfer_results}
\resizebox{63mm}{!}{
\begin{tabular}{c|c|cc|cc|cc}
\toprule[1pt]
\midrule
\multirow{2}{*}{Forgetting class \#}
  & 0 & \multicolumn{2}{c|}{100} & \multicolumn{2}{c|}{200} & \multicolumn{2}{c}{300}  \\ 
  & \multicolumn{1}{c|}{{\acc}}  & 

\multicolumn{1}{c|}{{\acc}}  & \multicolumn{1}{c|}{{\TIME}} &  
\multicolumn{1}{c|}{{\acc}}  & \multicolumn{1}{c|}{{\TIME}} & 
\multicolumn{1}{c|}{{\acc}}  & \multicolumn{1}{c}{{\TIME}} 
  \\

\midrule
\rowcolor{Gray}
\multicolumn{8}{c}{OxfordPets} \\
\midrule
Method \cite{jain2022data} 
 & \multirow{2}{*}{85.70}  & 85.79	& 71.84 &86.10 & 61.53  &86.32 & 54.53
 \\
 \MUSparse & & 85.83&35.47&	86.12&30.19& 86.26& 26.49
 \\
\midrule
\rowcolor{Gray}
\multicolumn{8}{c}{SUN397} \\
\midrule
 Method \cite{jain2022data}   & \multirow{2}{*}{46.55} & 46.97	& 73.26 &47.14& 61.43 &47.31 & 55.24
 				
 \\
 \MUSparse & & 47.20& 36.69 &	47.25& 30.96 &	47.37& 27.12	
 \\
\midrule
\bottomrule[1pt]
\end{tabular}
}
\vspace*{-8mm}
\end{wraptable}%
{\MUSparse} method achieves comparable or even slightly better 
transfer learning accuracy than the retraining-based approach \citep{jain2022data}.  Importantly, {\MUSparse} offers the advantage of computational efficiency 2$\times$ speed up over previous method \citep{jain2022data} across all cases, making it an appealing choice for transfer learning using large-scale models.
Here we remark that in order to align with previous method \cite{jain2022data}, we employed a fast-forward computer vision training pipeline  (FFCV) \citep{leclerc2022ffcv}
to accelerate our ImageNet training on GPUs.
\noindent \textbf{Additional results.} 
{
We found that model sparsity also enhances the privacy of the unlearned model, as evidenced by a lower {\MIAR}. Refer to Appendix\,\ref{appendix: additional results} and Fig.\,\ref{fig: results_privacy} for more results. In addition, we have expanded our experimental scope to encompass the `prune first, then unlearn' approach across various datasets and architectures. The results can be found in Tab.\,\ref{tab: overall_performance_ext_datasets}, Tab.\,\ref{tab: overall_performance_ext_archs}, and Tab.\,\ref{tab: overall_performance_ImageNet}. Furthermore, we conducted experiments on the $\ell_1$-sparse {\MU} across different datasets, the Swin-Transformer architecture, and varying model sizes within the ResNet family. The corresponding findings are presented in Fig.\,\ref{fig: results_l1_sparse_unlearn_others} and Tab.\,\ref{tab: sparse_MU vs MU}, \ref{tab: vit}, \ref{tab: overall_perfoamnce_arch_20} and \ref{tab: overall_perfoamnce_arch_50}.}

\section{Related Work}
\vspace*{-2mm}
While Sec.\,\ref{sec: primer_MU} provides a summary of related works concerning exact and approximate unlearning methods and metrics,  a more comprehensive review  is provided below.

\noindent \textbf{Machine unlearning.}
In addition to exact and approximate unlearning methods as we have reviewed in Sec.\,\ref{sec: primer_MU}, there  exists other literature aiming to  develop  the probabilistic notion of unlearning \cite{ginart2019making,guo2019certified,neel2021descent,ullah2021machine,sekhari2021remember}, in particular through the lens of differential privacy (DP) \cite{dwork2006our}. Although DP enables unlearning  with provable error guarantees, 
they typically require strong model and algorithmic assumptions and could lack effectiveness when facing practical  adversaries, \textit{e.g.}, membership inference attacks. Indeed, evaluating {\MU}  is far from trivial \cite{becker2022evaluating,thudi2022necessity,thudi2021unrolling}.  
Furthermore, the attention on  {\MU} has also   been raised in   different learning paradigms, \textit{e.g.}, federated learning   \cite{wang2022federated,liu2022right}, graph neural networks \cite{chen2022graph,chien2022certified,cheng2023gnndelete}, and adversarial ML \cite{marchant2022hard,di2022hidden}.
In addition to   preventing the leakage of  data privacy  from the trained models, the concept of {\MU} has also inspired  other emergent applications such as adversarial defense against backdoor attacks \cite{liu2022backdoor,warnecke2021machine} that we have studied and erasing image concepts of conditional generative models \cite{gandikota2023erasing,zhang2023forget}.

\noindent \textbf{Understanding data influence.}
The majority of {\MU} studies are motivated by data privacy. Yet, they  also closely relate to another line of research on understanding data influence in ML. For example, the influence function approach \cite{koh2017understanding} has been used as an algorithmic backbone of many unlearning methods  \cite{warnecke2021machine,izzo2021approximate}. From the viewpoint of data influence, {\MU}  has been used in the use case of adversarial defense against data poisoning backdoor attacks \cite{liu2022backdoor}. Beyond unlearning, evaluation of data influence  has also been studied in  fair learning  \cite{sattigeri2022fair,wang2022understanding},  transfer learning  \cite{jain2022data}, and   dataset pruning \cite{borsos2020coresets,yang2022dataset}. 



\noindent \textbf{Model pruning.}
The deployment constraints on \textit{e.g.}, computation, energy, and memory   necessitate the pruning of
today's ML models, \textit{i.e.}, promoting their weight sparsity. 
The vast majority of existing works \cite{han2015deep,chen2021lottery,frankle2018lottery,frankle2020linear,ma2021sanity,zhang2022advancing,blalock2020state} focus on  developing model pruning methods that can strike a graceful balance between model's generalization and sparsity.
In particular, the existence of LTH (lottery ticket hypothesis) \cite{frankle2018lottery} demonstrates 
the feasibility of co-improving the model's generalization  and efficiency (in terms of sparsity) \cite{liu2018rethinking,tanaka2020pruning,wang2020picking,lee2018snip,zhang2023data}. 
In addition to generalization, model sparsity   achieved by pruning   can also be  leveraged to improve other performance metrics, such as   robustness \cite{sehwag2020hydra,chen2022quarantine,diffenderfer2021winning}, model explanation  \cite{wong2021leveraging,chen2022can},
and privacy \cite{huang2020privacy,wang2020against,luo2021scalable,gong2020privacy}.

\section{Conclusion}
In this work, we advance the method of {machine unlearning} through a novel viewpoint: model sparsification, achieved by weight pruning. We show in both theory and practice that model sparsity plays a foundational and crucial role in closing the gap between exact unlearning and existing approximate unlearning methods. Inspired by that, we propose two new unlearning paradigms,  `prune first, then unlearn' and `sparsity-aware unlearn', which can significantly improve the efficacy of approximate unlearning. We demonstrate the effectiveness of our findings and proposals in extensive experiments across different unlearning setups. Our study also indicates the presence of \textit{model modularity} traits, such as weight sparsity, that could simplify the process of machine unlearning. This may open up exciting prospects for future research to investigate unlearning patterns within weight or architecture space.

\section{Acknowledgement}

{The work of J. Jia, J. Liu, Y. Yao, and S. Liu were supported by the Cisco Research Award and partially supported by the NSF Grant IIS-2207052, and the ARO Award W911NF2310343. Y. Liu was partially supported by NSF Grant IIS-2143895 and IIS-2040800. }

{{
\bibliographystyle{unsrtnat}
\bibliography{bibs/ref_JC_attack,bibs/ref_jh_model_dataset,bibs/ref_SL_pruning,bibs/unlearning}
}}
\newpage
\clearpage
\appendix
\onecolumn
\setcounter{section}{0}

\section*{Appendix}

\setcounter{section}{0}
\setcounter{figure}{0}
\makeatletter 
\renewcommand{\thefigure}{A\arabic{figure}}
\renewcommand{\theHfigure}{A\arabic{figure}}
\renewcommand{\thetable}{A\arabic{table}}
\renewcommand{\theHtable}{A\arabic{table}}

\makeatother
\setcounter{table}{0}

\setcounter{mylemma}{0}
\renewcommand{\themylemma}{A\arabic{mylemma}}
\setcounter{equation}{0}
\renewcommand{\theequation}{A\arabic{equation}}

\section{Proof of Proposition\,\ref{prop: IU}}
\label{appendix: IU}

Recap the definition of model update $ \Delta(\mathbf w) $ in \eqref{eq: Delta_IU} and $\thetafull = \btheta(\mathbf 1/N)$,
 we   approximate $ \Delta(\mathbf w) $ by the first-order Taylor expansion of $\btheta(\mathbf w)$ at $\mathbf w = \mathbf 1 / N$. This leads to
\begin{equation}
    \Delta(\mathbf w)  = \btheta(\mathbf w) - \btheta(\mathbf 1 / N) \approx \left. \frac{d \btheta(\mathbf w)}{d \mathbf w} \right|_{\mathbf w = \mathbf 1 / N} (\mathbf w - \mathbf 1 / N), \label{eq: Delta_IF}
\end{equation}
where $\frac{d \btheta(\mathbf w)}{d \mathbf w} \in \mathbb R^{M \times N}$, and recall that $M = |\thetafull|$ is the number of model parameters.  The gradient $\frac{d \btheta(\mathbf w)}{d \mathbf w}$ is known as implicit gradient \cite{gould2016differentiating} since it is defined through the solution of the optimization problem
  $\btheta(\mathbf w) = \argmin_{\btheta} L(\mathbf w, \btheta)$, where recall that $L(\mathbf w, \btheta) = \sum_{i=1}^N [w_i \ell_i (\btheta, \mathbf z_i)]$. By the stationary condition of $\btheta(\mathbf w)$, we obtain
\begin{equation}
    \nabla_{\btheta}L(\mathbf w, \btheta(\mathbf w)) = \mathbf 0. \label{eq: L_grad}
\end{equation}

Next, we take the derivative of  \eqref{eq: L_grad} w.r.t. $\mathbf w$ based on   the implicit function theorem \cite{gould2016differentiating} assuming that $\btheta(\mathbf w)$ is the  unique solution  to minimizing $L$. This leads to
\begin{equation}
     \left[\frac{d \btheta(\mathbf w)}{d \mathbf w}\right]^T \left[\left.\nabla_{\btheta, \btheta} L(\mathbf w, \btheta)\right|_{\btheta = \btheta(\mathbf w)} \right] +  \nabla_{\mathbf w, \btheta}L(\mathbf w, \btheta(\mathbf w)) = \mathbf 0,
\end{equation}
where $\nabla_{\mathbf a, \mathbf b} = \nabla_{\mathbf a} \nabla_{\mathbf b} \in \mathbb R^{|\mathbf a| \times |\mathbf b|}$ is the second-order partial derivative. Therefore,
\begin{equation}
    \frac{d \btheta(\mathbf w)}{d \mathbf w} = -\left[\nabla_{\btheta, \btheta} L(\mathbf w, \btheta(\mathbf w))\right]^{-1} \nabla_{\mathbf w, \btheta}L(\mathbf w, \btheta(\mathbf w))^T,
    \label{eq: IG}
\end{equation}
where $\nabla_{\mathbf w, \btheta}L(\mathbf w, \btheta(\mathbf w))$ can be expanded as
\begin{align}
    \nabla_{\mathbf w, \btheta}L(\mathbf w, \btheta(\mathbf w)) & = \nabla_{\mathbf w} \nabla_{\btheta} \sum_{i = 1}^{N}\left[w_i \ell_i(\btheta(\mathbf w), \mathbf z_i)\right]   \\ &= \nabla_{\mathbf w} \sum_{i = 1}^{N}\left[w_i \nabla_{\btheta} \ell_i(\btheta(\mathbf w), \mathbf z_i)\right]   \\ &= \left[
    \begin{array}{c}
    \nabla_{\btheta} \ell_1(\btheta(\mathbf w), \mathbf z_1)^T \\
    \nabla_{\btheta} \ell_2(\btheta(\mathbf w), \mathbf z_2)^T \\
    \vdots \\
    \nabla_{\btheta} \ell_N(\btheta(\mathbf w), \mathbf z_N)^T
    \end{array}\right].
    \label{eq: IG_cross2}
\end{align}

Based on \eqref{eq: IG} and \eqref{eq: IG_cross2}, we obtain the closed-form of implicit gradient at $\mathbf w = \mathbf 1/N$:
\begin{align}
    \frac{d \btheta(\mathbf w)}{d \mathbf w} \left. \right|_{\mathbf w = \mathbf 1 / N} = & -\left[\nabla_{\btheta, \btheta} L(\mathbf 1/N, \btheta(\mathbf 1/N))\right]^{-1} 
    \begin{bmatrix}
    \nabla_{\btheta} \ell_1(\btheta(\mathbf 1/N), \mathbf z_1) & \ldots &
    \nabla_{\btheta} \ell_N(\btheta(\mathbf 1/N), \mathbf z_N)
    \end{bmatrix} \nonumber \\
    =  & - \mathbf H^{-1}  \begin{bmatrix}
    \nabla_{\btheta} \ell_1(\btheta(\mathbf 1/N), \mathbf z_1) & \ldots &
    \nabla_{\btheta} \ell_N(\btheta(\mathbf 1/N), \mathbf z_N)
    \end{bmatrix},
    \label{eq: IG_2}
\end{align}
where $\mathbf H = \nabla_{\btheta, \btheta} L(\mathbf 1/N, \btheta(\mathbf 1/N))$.

Substituting \eqref{eq: IG_2} into  \eqref{eq: Delta_IF}, we obtain 
\begin{align}
\Delta(\mathbf w) & \approx 
- \mathbf H^{-1}  \begin{bmatrix}
    \nabla_{\btheta} \ell_1(\btheta(\mathbf 1/N), \mathbf z_1) & \ldots &
    \nabla_{\btheta} \ell_N(\btheta(\mathbf 1/N), \mathbf z_N)
    \end{bmatrix} (\mathbf w - \mathbf 1/N) \nonumber \\
    & = 
-\mathbf H^{-1} \sum_{i = 1}^{N} [ (w_i - 1 / N)\nabla_{\btheta} \ell_i(\btheta(\mathbf 1 / N), \mathbf z_i) ] \nonumber\\
& = \mathbf H^{-1} \nabla_{\btheta} L(\mathbf 1 / N - \mathbf w, \thetafull),
\end{align}
where the last equality holds by the definition of   $L(\mathbf w, \btheta) = \sum_{i=1}^N [w_i \ell_i (\btheta, \mathbf z_i)]$.

The proof is now complete.  \hfill $\square$

\textbf{Remark on {\IU} using ave-ERM vs. sum-ERM.}
Recall that the weighted empirical risk minimization (ERM) loss used in proposition\,\eqref{prop: IU}, $L(\mathbf w,\btheta) =  \sum_{i=1}^N [w_i \ell_i (\btheta, \mathbf z_i)]$   corresponds to the ave-ERM as $\mathbf w$ is subject to the simplex constraint ($\mathbf 1^T \mathbf w = 1$ and $\mathbf w \geq \mathbf 0$). This is different from the conventional derivation of {\IU} using  the sum-ERM \cite{koh2017understanding,guo2019certified} in the absence of simplex  constraint. In what follows, we discuss the impact of   ave-ERM on {\IU} vs. 
sum-ERM.

Noting that $\boldsymbol \theta_{\mathrm{o}}$ represents the original model trained through conventional ERM, namely, the weighted ERM loss by setting $w_i = c$ ($\forall i$) for a positive constant $c$.
Given the unlearning scheme (encoded in $\mathbf w$), the IU approach aims to delineate the model parameter adjustments required by MU from the initial model $\boldsymbol \theta_{\mathrm{o}}$. Such a model weight  modification is represented as $\Delta(\mathbf{w}) = \btheta(\mathbf w) - \btheta_{\mathrm{o}}$ mentioned in proposition\,\eqref{prop: IU}. The difference between ave-ERM and sum-ERM would play a role in  deriving $\Delta(\mathbf{w})$, which relies on the  {Taylor expansion} of $\btheta (\mathbf w)$ (viewed as a function of $\mathbf w$).
When the sum-ERM is considered, then the linearization point is typically set by $\mathbf w = \mathbf 1$. This leads to 
\begin{align}
  \Delta^{\mathrm{(sum)}}(\mathbf{w})&={\btheta}(\mathbf w) - {\btheta}(\mathbf 1)   \approx {\btheta}(\mathbf 1)+\frac{d\btheta(\mathbf w)}{d\mathbf w} \left | \right._{\mathbf w= \mathbf 1}  (\mathbf w-\mathbf 1) - {\btheta}(\mathbf 1) \nonumber \\
 & = \frac{d\btheta(\mathbf w)}{d\mathbf w}|_{\mathbf w= \mathbf 1} (\mathbf w-\mathbf 1),
 \label{eq: sum_erm}
\end{align}
where  $ \boldsymbol{\theta}(\mathbf 1) = \btheta_{\mathrm{o}}$  for sum-ERM, and
$\frac{d\btheta(\mathbf w)}{d\mathbf w}$ is  implicit gradient \cite{gould2016differentiating} since it is defined upon an implicit optimization problem ${\btheta}(\mathbf w) = \arg\min_{\btheta} L(\mathbf w, \btheta)$.


When the ave-ERM is considered, the linearization point is set by $\mathbf w = \mathbf 1/N$. This leads to 
\begin{align}
  \Delta^{\mathrm{(ave)}}(\mathbf{w})&={\btheta}(\mathbf w) - {\btheta}(\mathbf 1/N)   \approx {\btheta}(\mathbf 1/N)+\frac{d\btheta(\mathbf w)}{d\mathbf w} |_{\mathbf w= \mathbf 1/N}  (\mathbf w-\mathbf 1/N) - {\btheta}(\mathbf 1/N) \nonumber \\
 & = \frac{d\btheta(\mathbf w)}{d\mathbf w}|_{\mathbf w= \mathbf 1/N} (\mathbf w-\mathbf 1/N),
 \label{eq: ave_erm}
\end{align}
where   $   \boldsymbol{\theta}(\mathbf 1/N) = \boldsymbol \theta_{\mathrm{o}}$ for ave-ERM. The derivation of the implicit gradient $\frac{d\btheta(\mathbf w)}{d\mathbf w}$ is shown in \eqref{eq: IG_2}.


Next, let us draw a  comparison between $\Delta^{\mathrm{(sum)}}(\mathbf{w})$ and $\Delta^{\mathrm{(ave)}}(\mathbf{w})$ using a specific example below. 


If we aim to unlearn the first $k$ training data points, the unlearning weights $\mathbf w_{\mathrm{MU}}$ under sum-ERM is then given by  $\mathbf w_{\mathrm{MU}}^{\mathrm{(sum)}} = [\underbrace{0, 0, \ldots, 0}_{k~ 0s}, 1, 1,\ldots, 1]$, where $0$ encodes the data sample to be unlearned or removed. This yields $(\mathbf w_{\mathrm{MU}}^{\mathrm{(sum)}} - \mathbf 1) = [\underbrace{1, 1, \ldots, 1}_{k ~ 1s}, 0, 0, \ldots, 0]$. By contrast, the unlearning weights $\mathbf w_{\mathrm{MU}}$ under ave-ERM is   given by  $\mathbf w_{\mathrm{MU}}^{\mathrm{(ave)}} = [\underbrace{0, 0, \ldots, 0}_{k~ 0s}, \frac{1}{N-k}, \frac{1}{N-k},\ldots, \frac{1}{N-k}]$. As a result, $(\mathbf w_{\mathrm{MU}}^{\mathrm{(ave)}}-\mathbf 1/N) = [\underbrace{-\frac{1}{N}, -\frac{1}{N}, \ldots, -\frac{1}{N}}_{k~ \frac{1}{N}s}, \frac{1}{N-k} - \frac{1}{N}, \frac{1}{N-k}- \frac{1}{N},\ldots, \frac{1}{N-k}- \frac{1}{N}]$. The above difference is caused by the presence of simplex constraint of $\mathbf w$ in ave-ERM. Thus, the MU's weight configuration $(\mathbf w_{\mathrm{MU}}^{\mathrm{(ave)}}-\mathbf 1/N)$ obtained from ave-ERM is different from $(\mathbf w_{\mathrm{MU}}^{\mathrm{(sum)}} - \mathbf 1)$ in the sum-ERM setting. 

Given the above example, the error term of the Taylor expansion using sum-ERM for $\mathbf w = \mathbf w_{\mathrm{MU}}^{\mathrm{sum}}$ is in the order of $\|\mathbf w_{\mathrm{MU}}^{\mathrm{sum}} - \mathbf 1\|_2^2=k$, while the error term using ave-ERM for $\mathbf w = \mathbf w_{\mathrm{MU}}^{\mathrm{ave}}$ is in the order of $\|\mathbf w_{\mathrm{MU}}^{\mathrm{ave}} - \mathbf 1/N\|_2^2=\frac{k}{N^2} + \frac{k^2}{N^2(N-k)} = \frac{k}{N(N-k)}$. Thus compared to ave-ERM, the use of sum-ERM could cause the first-order Taylor expansion in IU less accurate as the number of unlearning datapoints ($k$) increases. Furthermore, the IG $\frac{d\boldsymbol\theta(\mathbf w)}{d\mathbf w}|_{\mathbf w= \mathbf 1}$ in sum-ERM is also different from $\frac{d\boldsymbol\theta(\mathbf w)}{d\mathbf w}|_{\mathbf w= \mathbf 1/N}$ in ave-ERM as they are evaluated at two different linearization points.

\section{Proof of Proposition\,\ref{prop: SGD_sparse_MU}}
\label{appendix: SGD_sparse_MU}

{The proof follows  \cite[Sec.\,5]{thudi2021unrolling}, with the additional condition that the model is \textbf{sparse} encoded by a pre-fixed (binary) pruning mask $\mathbf m$, namely,
$\btheta^\prime \Def \mathbf m \odot \btheta$.} 
Then, based on \cite[Eq.\,5]{thudi2021unrolling}, the model updated by SGD yields  
\begin{align}
    \btheta_t^\prime \approx  \btheta_0^\prime - \eta \mathbf m \odot \sum_{i=1}^{t-1}\nabla_{\btheta} \ell (\btheta_0^\prime, \hat{\mathbf z}_i) + 
    \mathbf m \odot (\sum_{i=1}^{t-1} f(i)),
  \label{eq: GA_unroll_SGD}
 \end{align} 
where $\btheta_0^\prime = \mathbf m \odot \btheta_0$ is the  model initialization when using SGD-based sparse training,    $\{ \hat{\mathbf z}_i \}$
is the sequence of stochastic data samples, $t$ is the number of training iterations, $\eta$ is the learning rate,  and 
 $f(i)$ is defined recursively as 
\begin{align}
    f(i) = -\eta \nabla^2_{\btheta, \btheta} \ell (\btheta_0^\prime, \hat{\mathbf z}_i) \left( -\eta \sum_{j=0}^{i-1} {{ \mathbf m \odot \nabla_{\btheta} \ell (\btheta_0^\prime, \hat{\mathbf z}_j) }} + \sum_{j=0}^{i-1} {{( \mathbf m \odot f(j))}} \right),
    \label{eq: f_i_unroll_SGD}
\end{align}
with $f(0) = 0$. 
Inspired by the second term of \eqref{eq: GA_unroll_SGD}, to unlearn the data sample $\hat{\mathbf z}_i$, we will have   to   add back the first-order gradients under $\hat{\mathbf z}_i$. This corresponds to the {\GA}-based approximate unlearning method. Yet, this approximate unlearning introduces an unlearning error, given by the last term of \eqref{eq: GA_unroll_SGD}
\begin{align}
\mathbf e_{\mathbf m}(  \btheta_0, \{ \hat{\mathbf z}_i \},  t, \eta) \Def       \mathbf m \odot (\sum_{i=1}^{t-1} f(i)). 
     \label{eq: unlearn_err_sparse}
\end{align}

Next, if we interpret the mask $\mathbf m$ as a diagonal matrix $\mathrm{diag}(\mathbf m)$ with $0$'s and $1$'s along its diagonal based on $\mathbf m$, we can then express the sparse model $\mathbf m \odot \btheta$ as $\mathrm{diag}(\mathbf m) \btheta$.
Similar to  \cite[Eq.\,9]{thudi2021unrolling}, we can derive a bound on the unlearning error \eqref{eq: unlearn_err_sparse} by ignoring the terms other than those with $\eta^2$ in $f(i)$, \textit{i.e.}, \eqref{eq: f_i_unroll_SGD}. 
This is because, in the recursive form of $f(i)$, all other terms exhibit a higher degree of the learning rate $\eta$ compared to $\eta^2$. As a result, we obtain
  \begin{align}
  &  e(\mathbf m) =   \left\|  \mathbf e_{\mathbf m}(\btheta_0, \{ \hat{\mathbf z}_i \},  t, \eta) \right\|_2 = \left\| \mathbf m \odot (\sum_{i=1}^{t-1} f(i)) \right\|_2 \nonumber \\
     & \approx \eta^2 \left\| \mathrm{diag}(\mathbf m)   \sum_{i=1}^{t-1} \nabla^2_{\btheta, \btheta} \ell (\btheta_0^\prime, \hat{\mathbf z}_i) \sum_{j=0}^{i-1} \mathbf m \odot \nabla_{\btheta} \ell (\btheta_0^\prime, \hat{\mathbf z}_j) \right\|_2 \nonumber \\ 
     &\leq \eta^2 \sum_{i=1}^{t-1} \left\| \mathrm{diag}(\mathbf m)  \nabla^2_{\btheta, \btheta} \ell (\btheta_0^\prime, \hat{\mathbf z}_i) \sum_{j=0}^{i-1} \mathbf m \odot \nabla_{\btheta} \ell (\btheta_0^\prime, \hat{\mathbf z}_j) \right\|_2 \tag{Triangle inequality} \\
     & \leq \eta^2 \sum_{i=1}^{t-1} \left\| \mathrm{diag}(\mathbf m) 
    \nabla^2_{\btheta, \btheta} \ell (\btheta_0^\prime, \hat{\mathbf z}_i) \right\| \left\| \sum_{j=0}^{i-1} \mathbf m \odot \nabla_{\btheta} \ell (\btheta_0^\prime, \hat{\mathbf z}_j) \right\|_2 \label{eq: third_ineq_bound} \\ 
    &\lesssim \eta^2 \sum_{i=1}^{t-1} \left\| \mathrm{diag}(\mathbf m)  \nabla^2_{\btheta, \btheta} \ell (\btheta_0^\prime, \hat{\mathbf z}_i) \right\| \frac{i}{t } \left\|\btheta_t^\prime - \btheta_0^\prime \right\|_2  \label{eq: second_ineq_bound} \\
     & \leq \eta^2  \sigma (\mathbf m)   
     \left\| \mathbf m \odot (\btheta_t - \btheta_0 )\right\|_2  \frac{1}{t } \frac{t-1}{2} t = \frac{\eta^2}{2}{(t-1) \| 
   \mathbf m \odot (\btheta_t - \btheta_0) \|_2}  \sigma(\mathbf m),
   \label{eq: norm_sparse_MU_err}
 \end{align}
where the   inequality \eqref{eq: second_ineq_bound} holds given the fact that  
$\sum_{j=0}^{i-1} \mathbf m \odot \nabla_{\btheta} \ell (\btheta_0^\prime, \hat{\mathbf z}_j)$ in \eqref{eq: third_ineq_bound} can be approximated by its expectation $\frac{i (\btheta_t^\prime - \btheta_0^\prime)}{t}$  \cite[Eq.\,7]{thudi2021unrolling}, and
$\sigma(\mathbf m) \Def \max_{j} \{  \sigma_{j}( \nabla_{\btheta,\btheta}^2\ell ), \text{if } m_j \neq 0  \}$, \textit{i.e.}, the largest eigenvalue among the dimensions left intact by the binary mask $\mathbf m$. 
The above suggests that the unlearning error might be large if $\mathbf m = \mathbf 1$ (no pruning). 
Based on \eqref{eq: norm_sparse_MU_err}, we can then readily obtain the big $O$ notation in \eqref{eq: err_bd_SGD_sparse}.
This completes the proof.

\section{Additional Experimental Details and Results}
\label{appendix: additional results and details}

\subsection{Datasets and models}
\label{appendix: datasets and models}

We summarize the datasets and model configurations in Tab.\,\ref{tab: dataset_model_settings}.  
\begin{table*}[htb]
\centering
\caption{Dataset and model setups. }
\resizebox{0.6\textwidth}{!}{
\begin{tabular}{c|c|c|c|c|c}
\toprule[1pt]
\midrule
\multirow{2}{*}{Settings} & \multicolumn{2}{c|}{CIFAR-10}    & SVHN      & CIFAR-100 & ImageNet  \\

                          & ResNet-18 & VGG-16 & ResNet-18 & ResNet-18 & ResNet-18        \\
\midrule
Batch Size                & 256             & 256    & 256       & 256     & 1024          \\
    \midrule
\bottomrule
\end{tabular}}
\label{tab: dataset_model_settings}
\end{table*}

\subsection{Additional training and unlearning settings}
\label{appendix: training and unlearning settings}

\noindent \textbf{Training configuration of pruning.} For all pruning methods, including IMP \cite{frankle2018lottery}, SynFlow \cite{tanaka2020pruning}, and OMP \cite{ma2021sanity}, we adopt the settings from the current SOTA implementations \cite{ma2021sanity}; see a summary in Tab.\,\ref{tab: pruning_settings}. For IMP, OMP, and SynFlow, we adopt the step learning rate scheduler with a decay rate of 0.1 at $50\%$ and $75\%$ epochs. We adopt $0.1$ as the initial learning rate for all pruning methods. 
\begin{table*}
    \centering
        \caption{Detailed training details for model pruning.}
    \resizebox{0.50\textwidth}{!}{
    \begin{tabular}{c|c|c|c}
    
    \toprule[1pt]
    \midrule
    Experiments & CIFAR-10/CIFAR-100 & SVHN & ImageNet  \\
    \midrule
    Training epochs & 182 & 160 & 90 \\
    \midrule
    Rewinding epochs & 8&8 & 5  \\
    \midrule
    Momentum & 0.9 & 0.9 & 0.875 \\
    \midrule
    $\ell_2$ regularization & $5e^{-4}$ & $5e^{-4}$ & $3.05e^{-5}$  \\
    \midrule
    Warm-up epochs & 1(75 for VGG-16)&0 &8\\
    \midrule
    \bottomrule
    \end{tabular}
    }

\label{tab: pruning_settings}
\end{table*}

\noindent \textbf{Additional training details of {\MU}.}
For all datasets and model architectures, we adopt $10$ epochs for {\FT}, and $5$ epochs for {\GA} method. The learning rate for {\FT} and {\GA} are carefully tuned between $[10^{-5},0.1]$ for each dataset and model architecture. In particular, we adopt $0.01$ as the learning rate for {\FT} method and $10^{-4}$ for {\GA} on the CIFAR-10 dataset (ResNet-18, class-wise forgetting) at different sparsity levels.  By default, we choose SGD as the optimizer for the {\FT} and {\GA} methods. As for {\FF} method, we perform a greedy search for  hyperparameter tuning \cite{golatkar2020eternal} between $10^{-9}$ and ${10^{-6}}$. 

\subsection{Detailed metric settings}
\label{appendix: metric settings}
\noindent \textbf{Details of MIA implementation.}
MIA is implemented using the prediction confidence-based attack method \cite{song2020systematic}. There are mainly two phases during its computation: \textbf{(1) training phase}, and \textbf{(2) testing phase}.  
To train an MIA model, we first sample a balanced dataset from the remaining dataset ($\Dr$) and the test dataset (different from the forgetting dataset $\Df$) to train the MIA predictor. The learned MIA is then used for MU evaluation in its testing phase. 
To evaluate the performance of {\MU}, {\MIAF} is obtained by applying the  learned MIA predictor to the unlearned model ($\thetaunl$) on the forgetting dataset ($\Df$). Our objective is to find out how many samples in $\Df$ can be correctly predicted as non-training samples by the MIA model against $\thetaunl$. The formal definition of {\MIAF} is then given by:
\begin{equation}
\text{\MIAF} = \frac{TN}{|\Df|},
\end{equation}
where $TN$ refers to the true negatives predicted by our MIA predictor, \textit{i.e.}, the number of the forgetting samples predicted as non-training examples, and $|\Df|$ refers to the size of the forgetting dataset.
As described above, {\MIAF} leverages the privacy attack to justify how good the unlearning performance could be.

\subsection{Additional experiment results}
\label{appendix: additional results}



\noindent \textbf{Model sparsity benefits privacy of {\MU} for `free'.}
It was  recently shown in \cite{huang2020privacy,wang2020against} that model sparsification helps protect data privacy, in terms of defense against MIA  used to infer training data  information   from a learned model. Inspired by the above, we ask if sparsity can also bring  the privacy benefit to an unlearned model, evaluated by  the MIA rate on the remaining dataset $\Dr$ (that we term \textbf{\MIAR}).  This is different from  {\MIAF}, which
reflects the efficacy of scrubbing ${\Df}$, \textit{i.e.}, correctly predicting that    data sample in ${\Df}$ is not in the training set of the unlearned model. In contrast, 
{\MIAR}  characterizes the   \textit{privacy}   of the unlearned model about $\Dr$. A \textit{lower} {\MIAR}   implies \textit{less} information leakage.


\begin{wrapfigure}{r}{40mm}
\centerline{
\includegraphics[width=40mm]{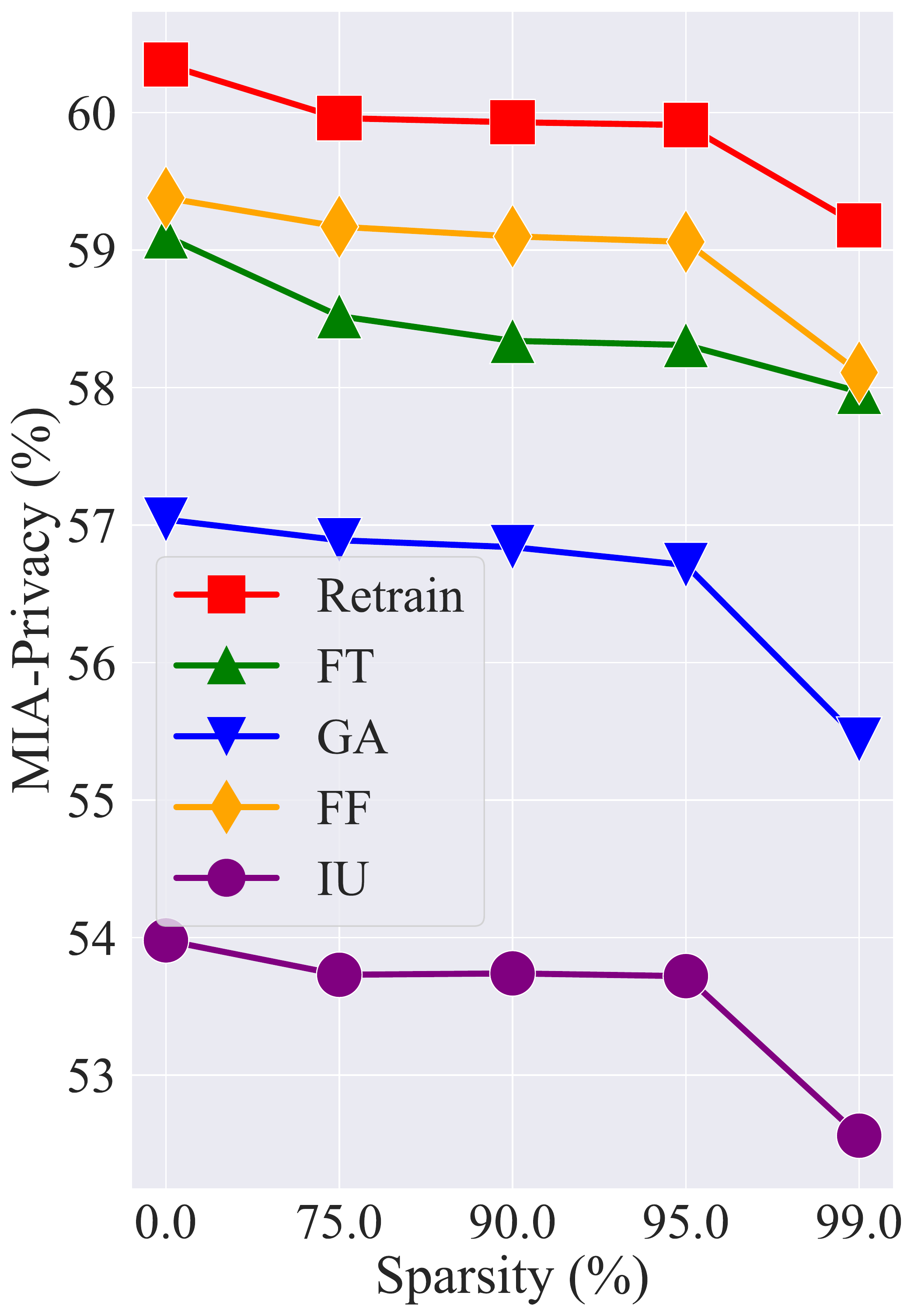}
}
\vspace*{-1.5mm}
\caption{\footnotesize{Privacy on $\Dr$ ({\MIAR}) using different unlearning methods vs. model sparsity.
}}
  \label{fig: results_privacy}
\vspace*{-6.5mm}
  
\end{wrapfigure}
\textbf{Fig.\,\ref{fig: results_privacy}} shows {\MIAR} of
unlearned models versus the   sparsity ratio applied to different   unlearning methods in the `prune first, then unlearn' paradigm. 
As we can see, 
{\MIAR} decreases as the sparsity   increases. This suggests the improved  privacy 
of  unlearning on sparse models.
Moreover, we observe that approximate unlearning outperforms exact unlearning ({\retrain}) in   privacy preservation of $\Dr$. This is because {\retrain}  is conducted over ${\Dr}$ from scratch, leading to the strongest dependence on ${\Dr}$ than other unlearning methods.  Another interesting observation is that {\IU} and {\GA} yield a much smaller  {\MIAR}  than other approximate unlearning methods. The rationale behind that is   {\IU} and {\GA} have a weaker correlation with ${\Dr}$ during unlearning. Specifically,  the unlearning loss of {\IU}  only involves the forgetting data influence weights, \textit{i.e.}, $(\mathbf 1/N - \mathbf w)$ in \eqref{eq: Delta_IU}.
Similarly, {\GA} only performs gradient ascent   over $\Df$, with the least dependence on $\Dr$.

\noindent \textbf{Performance of `prune first, then unlearn' on various datasets and architectures.}
As demonstrated in Tab.\,\ref{tab: overall_performance_ext_datasets} and Tab.\,\ref{tab: overall_performance_ext_archs}, the introduction of model sparsity can effectively reduce the discrepancy between approximate and exact unlearning across a diverse range of datasets and architectures. This phenomenon is observable in various unlearning scenarios. Remarkably, model sparsity enhances both {\UA} and {\MIAF} metrics without incurring substantial degradation on {\RA} and {\TA} in different unlearning scenarios. These observations corroborate the findings reported in Tab.\,\ref{tab: overall_performance}. 

\begin{table*}[htb]
\centering
\caption{MU  performance vs. sparsity on additional datasets (CIFAR-100 \cite{krizhevsky2009learning} and SVHN \cite{netzer2011reading}) for both class-wise forgetting and random data forgetting. The content format follows Tab.\,\ref{tab: overall_performance}.
}
\label{tab: overall_performance_ext_datasets}
\resizebox{0.95\textwidth}{!}{
\begin{tabular}{c|cc|cc|cc|cc|c}
\toprule[1pt]
\midrule
  \multirow{2}{*}{\MU}& \multicolumn{2}{c|}{{\UA}} & \multicolumn{2}{c|}{{\MIAF}}& \multicolumn{2}{c|}{{\RA}} & \multicolumn{2}{c|}{{\TA}}&{\RTE}  \\ 
  & \multicolumn{1}{c|}{{\textsc{Dense}}}  & \multicolumn{1}{c|}{$\mathbf{95\%}$ \textbf{Sparsity}}
    & \multicolumn{1}{c|}{\textsc{Dense}}  & \multicolumn{1}{c|}{$\mathbf{95\%}$ \textbf{Sparsity}}
    & \multicolumn{1}{c|}{\textsc{Dense}}  & \multicolumn{1}{c|}{$\mathbf{95\%}$ \textbf{Sparsity}}
      & \multicolumn{1}{c|}{\textsc{Dense}}  & \multicolumn{1}{c|}{$\mathbf{95\%}$ \textbf{Sparsity}} & (min)
  \\

\midrule
\rowcolor{Gray}
\multicolumn{10}{c}{Class-wise forgetting, CIFAR-100} \\
\midrule
 \retrain &\textcolor{blue}{$100.00_{\pm{0.00}}$}   & \textcolor{blue}{$100.00_{\pm{0.00}}$} 
  &\textcolor{blue}{$100.00_{\pm{0.00}}$}   & \textcolor{blue}{$100.00_{\pm{0.00}}$}
   &\textcolor{blue}{$99.97_{\pm{0.01}}$}   & \textcolor{blue}{$96.68_{\pm{0.15}}$}
   &\textcolor{blue}{$73.74_{\pm{0.19}}$}   & \textcolor{blue}{$69.49_{\pm{0.41}}$} 
& $48.45$
 \\
 \FT &$15.00_{\pm{4.86}}$ (\textcolor{blue}{$85.00$})      & ${46.39}_{\pm{5.59}}$ (\textcolor{blue}{$53.61$}) 
  &$66.11_{\pm{4.53}}$ (\textcolor{blue}{$33.89$})    & ${67.17}_{\pm{5.91}}$ (\textcolor{blue}{$32.83$}) 
   &$99.87_{\pm{0.05}}$(\textcolor{blue}{$0.10$})   & $97.78_{\pm{0.52}}$ (\textcolor{blue}{$1.10$}) 
   &$74.88_{\pm{0.18}}$ (\textcolor{blue}{$1.14$}) &  ${72.53}_{\pm{0.11}}$ (\textcolor{blue}{$3.04$}) &$1.90$
 \\
 \GA  &$81.47_{\pm{0.32}}$(\textcolor{blue}{$18.53$})   & ${99.01}_{\pm{0.01}}$ (\textcolor{blue}{$0.99$}) 
 & $93.47_{\pm{4.56}}$ (\textcolor{blue}{$6.53$})  &${100.00}_{\pm{0.00}}$ (\textcolor{blue}{$0.00$}) 
 & $90.33_{\pm{1.71}}$ (\textcolor{blue}{$9.64$}) &$80.45_{\pm{0.78}}$ (\textcolor{blue}{$16.23$}) 
 & $64.94_{\pm{0.74}}$ (\textcolor{blue}{$8.80$})  & $60.99_{\pm{0.14}}$ (\textcolor{blue}{$8.50$})  
& $0.21$
 \\
\IU &$84.12_{\pm{0.34}}$ (\textcolor{blue}{$15.88$}) &${99.78}_{\pm{0.01}}$ (\textcolor{blue}{$0.22$}) 
& $98.44_{\pm{0.45}}$ (\textcolor{blue}{$1.56$}) &${99.33}_{\pm{0.00}}$  (\textcolor{blue}{$0.67$}) 
& $96.23_{\pm{0.02}}$ (\textcolor{blue}{$3.74$}) & $95.45_{\pm{0.17}}$ (\textcolor{blue}{$1.23$}) 
&$71.24_{\pm{0.22}}$ (\textcolor{blue}{$2.50$}) & ${70.79}_{\pm{0.11}}$ (\textcolor{blue}{$0.95$}) 
& $4.30$
\\
\midrule
\rowcolor{Gray}
\multicolumn{10}{c}{Random data forgetting, CIFAR-100} \\
\midrule
 \retrain &\textcolor{blue}{$24.76_{\pm{0.12}}$}   & \textcolor{blue}{$27.64_{\pm{1.03}}$} 
  &\textcolor{blue}{$49.80_{\pm{0.26}}$}   & \textcolor{blue}{$44.87_{\pm{0.81}}$}
   &\textcolor{blue}{$99.98_{\pm{0.02}}$}   & \textcolor{blue}{$99.24_{\pm{0.02}}$}
   &\textcolor{blue}{$74.46_{\pm{0.08}}$}   & \textcolor{blue}{$69.78_{\pm{0.15}}$} 
& $48.70$
 \\
 
 \FT &$0.78_{\pm{0.34}}$ (\textcolor{blue}{$23.98$})    & ${8.37}_{\pm{1.63}}$ (\textcolor{blue}{$19.27$}) 
  & $11.13_{\pm{0.40}}$ (\textcolor{blue}{$38.67$})   & ${18.57}_{\pm{1.57}}$ (\textcolor{blue}{$26.30$}) 
   &$99.93_{\pm{0.02}}$ (\textcolor{blue}{$0.05$})   & $99.20_{\pm{0.27}}$ (\textcolor{blue}{$0.04$}) 
   &$75.14_{\pm{0.09}}$ (\textcolor{blue}{$0.68$}) &  ${73.18}_{\pm{0.30}}$ (\textcolor{blue}{$1.60$}) &$3.74$
 \\
 \GA  &$0.04_{\pm{0.02}}$(\textcolor{blue}{$24.75$})   & ${3.92}_{\pm{0.28}}$ (\textcolor{blue}{$23.72$}) 
 & $3.80_{\pm{0.87}}$ (\textcolor{blue}{$46.00$})  &${7.51}_{\pm{1.37}}$ (\textcolor{blue}{$37.36$}) 
 & $99.97_{\pm{0.01}}$ (\textcolor{blue}{$0.01$}) &$98.40_{\pm{1.22}}$ (\textcolor{blue}{$0.84$}) 
 & $74.07_{\pm{0.11}}$ (\textcolor{blue}{$0.39$})  & $72.19_{\pm{0.15}}$ (\textcolor{blue}{$2.41$})  
& $0.24$
 \\
\IU &$1.53_{\pm{0.36}}$ (\textcolor{blue}{$23.23$}) &${6.01}_{\pm{0.17}}$ (\textcolor{blue}{$21.63$}) 
& $6.58_{\pm{0.42}}$ (\textcolor{blue}{$43.22$}) &${11.47}_{\pm{0.54}}$  (\textcolor{blue}{$33.40$}) 
& $99.01_{\pm{0.28}}$ (\textcolor{blue}{$0.97$}) & $96.53_{\pm{0.24}}$ (\textcolor{blue}{$2.71$}) 
&$71.76_{\pm{0.31}}$ (\textcolor{blue}{$2.70$}) & ${69.40}_{\pm{0.19}}$ (\textcolor{blue}{$0.38$}) 
& $3.80$
\\
\midrule
\rowcolor{Gray}
\multicolumn{10}{c}{Class-wise forgetting, SVHN} \\
\midrule
 \retrain &\textcolor{blue}{$100.00_{\pm{0.00}}$}  & \textcolor{blue}{$100.00_{\pm{0.00}}$} 
 & \textcolor{blue}{$100.00_{\pm{0.00}}$}&  \textcolor{blue}{$100.00_{\pm{0.00}}$ }
& \textcolor{blue}{$100.00_{\pm{0.00}}$} &  \textcolor{blue}{$100.00_{\pm{0.00}}$}  
 &  \textcolor{blue}{$95.71_{\pm{0.12}}$}& \textcolor{blue}{$94.95_{\pm{0.05}}$ } & $42.84$
 \\
 \FT & {$11.48_{\pm{8.12}}$ } (\textcolor{blue}{$88.52$})    & ${51.93}_{\pm{19.62}}$ (\textcolor{blue}{$48.07$}) 
& $86.12_{\pm{9.62}}$ (\textcolor{blue}{$13.88$}) &${99.42}_{\pm{0.51}}$  (\textcolor{blue}{$0.58$}) 
& $100.00_{\pm{0.00}}$ (\textcolor{blue}{$0.00$}) &  $99.00_{\pm{0.00}}$ (\textcolor{blue}{$1.00$}) 
& $95.99_{\pm{0.07}}$ (\textcolor{blue}{$0.28$})&  $95.89_{\pm{0.02}}$ (\textcolor{blue}{$0.94$}) & $2.86$
 \\
  \GA &$83.87_{\pm{0.19}}$ (\textcolor{blue}{$16.13$})   & ${86.52}_{\pm{0.11}}$ (\textcolor{blue}{$13.48$}) 
  & $99.97_{\pm{0.02}}$ (\textcolor{blue}{$0.03$})&  ${100.00}_{\pm{0.00}}$ (\textcolor{blue}{$0.00$}) 
  & $99.60_{\pm{0.15}}$ (\textcolor{blue}{$0.40$})&$98.37_{\pm{0.11}}$ (\textcolor{blue}{$1.63$})  
  &  $95.27_{\pm{0.02}}$ (\textcolor{blue}{$0.44$})& $93.42_{\pm{0.07}}$ (\textcolor{blue}{$1.53$}) & $0.28$
  
  \\
\IU &$95.11_{\pm{0.02}}$ (\textcolor{blue}{$4.89$})   & ${100.00}_{\pm{0.00}}$ (\textcolor{blue}{$0.00$}) 
&$99.89_{\pm{0.04}}$ (\textcolor{blue}{$0.11$})  & ${100.00}_{\pm{0.00}}$(\textcolor{blue}{$0.00$}) 
&$100.00_{\pm{0.00}}$ (\textcolor{blue}{$0.00$})   & $99.85_{\pm{0.02}}$ (\textcolor{blue}{$0.15$}) 
&$95.70_{\pm{0.09}}$ (\textcolor{blue}{$0.01$})   & $94.90_{\pm{0.04}}$ (\textcolor{blue}{$0.05$}) & $3.19$
\\
\midrule
\rowcolor{Gray}
\multicolumn{10}{c}{Random data forgetting, SVHN} \\
\midrule
 \retrain &\textcolor{blue}{$4.89_{\pm{0.11}}$}&\textcolor{blue}{$ 4.78_{\pm{0.23}}$}&\textcolor{blue}{$15.38_{\pm{0.14}}$}&\textcolor{blue}{$15.25_{\pm{0.18}}$}&\textcolor{blue}{$100.00_{\pm{0.00}}$}&\textcolor{blue}{$100.00_{\pm{0.00}}$}&\textcolor{blue}{$95.54_{\pm{0.09}}$}&\textcolor{blue}{$95.44_{\pm{0.12}}$} & 42.71
\\
 \FT & $2.28_{\pm{1.41}}$ (\textcolor{blue}{$2.61$})& $3.77_{\pm{0.13}}$ (\textcolor{blue}{$1.01$})& $6.14_{\pm{3.30}}$ (\textcolor{blue}{$9.24$})& $8.38_{\pm{0.42}}$ (\textcolor{blue}{$6.87$})& $99.71_{\pm{0.33}}$ (\textcolor{blue}{$0.29$})& $99.31_{\pm{0.33}}$ (\textcolor{blue}{$0.69$})& $94.77_{\pm{0.87}}$ (\textcolor{blue}{$0.77$})& $93.92_{\pm{0.34}}$ (\textcolor{blue}{$1.52$}) & 2.73
 \\
 \GA & $0.99_{\pm{0.42}}$ (\textcolor{blue}{$3.90$})& $2.68_{\pm{0.53}}$ (\textcolor{blue}{$2.10$})& $3.07_{\pm{0.53}}$ (\textcolor{blue}{$12.31$})& $9.31_{\pm{0.48}}$ (\textcolor{blue}{$5.94$})& $99.43_{\pm{0.22}}$ (\textcolor{blue}{$0.57$})& $97.83_{\pm{0.43}}$ (\textcolor{blue}{$2.17$})& $94.03_{\pm{0.21}}$ (\textcolor{blue}{$1.51$})& $93.33_{\pm{0.27}}$ (\textcolor{blue}{$2.11$}) & 0.26

  \\
\IU &$3.48_{\pm{0.13}}$ (\textcolor{blue}{$1.41$})   & ${5.62}_{\pm{0.48}}$ (\textcolor{blue}{$0.84$}) 
&$9.44_{\pm{0.27}}$ (\textcolor{blue}{$5.94$})  & ${12.28}_{\pm{0.41}}$(\textcolor{blue}{$2.97$}) 
&$96.30_{\pm{0.08}}$ (\textcolor{blue}{$3.70$})   & $95.67_{\pm{0.15}}$ (\textcolor{blue}{$4.33$}) 
&$91.59_{\pm{0.11}}$ (\textcolor{blue}{$3.95$})   & $90.91_{\pm{0.26}}$ (\textcolor{blue}{$4.53$}) & $3.21$
\\
\midrule
\bottomrule[1pt]
\end{tabular}
}

\end{table*}

\begin{table*}[htb]
\centering
\caption{MU performance vs. sparsity on the additional architecture (VGG-16 \cite{simonyan2014very}) for both class-wise forgetting and random data forgetting on CIFAR-10.  The content format follows Tab.\,\ref{tab: overall_performance}.
}
\label{tab: overall_performance_ext_archs}
\resizebox{0.95\textwidth}{!}{
\begin{tabular}{c|cc|cc|cc|cc|c}
\toprule[1pt]
\midrule
  \multirow{2}{*}{\MU}& \multicolumn{2}{c|}{{\UA}} & \multicolumn{2}{c|}{{\MIAF}}& \multicolumn{2}{c|}{{\RA}} & \multicolumn{2}{c|}{{\TA}}&{\RTE}  \\ 
  & \multicolumn{1}{c|}{{\textsc{Dense}}}  & \multicolumn{1}{c|}{$\mathbf{95\%}$ \textbf{Sparsity}}
    & \multicolumn{1}{c|}{\textsc{Dense}}  & \multicolumn{1}{c|}{$\mathbf{95\%}$ \textbf{Sparsity}}
    & \multicolumn{1}{c|}{\textsc{Dense}}  & \multicolumn{1}{c|}{$\mathbf{95\%}$ \textbf{Sparsity}}
      & \multicolumn{1}{c|}{\textsc{Dense}}  & \multicolumn{1}{c|}{$\mathbf{95\%}$ \textbf{Sparsity}} & (min)
  \\

\midrule
\rowcolor{Gray}
\multicolumn{10}{c}{Class-wise forgetting, VGG-16} \\
\midrule
 \retrain &\textcolor{blue}{$100.00_{\pm{0.00}}$}   & \textcolor{blue}{$100.00_{\pm{0.00}}$} 
  &\textcolor{blue}{$100.00_{\pm{0.00}}$} &  \textcolor{blue}{$100.00_{\pm{0.00}}$}
   &\textcolor{blue}{$100.00_{\pm{0.01}}$}  & \textcolor{blue}{$99.97_{\pm{0.00}}$}
   &\textcolor{blue}{$94.83_{\pm{0.10}}$}   & \textcolor{blue}{$92.93_{\pm{0.06}}$} & $30.38$

 \\
 \FT &$28.00_{\pm{8.16}}$ (\textcolor{blue}{$72.00$})      & ${34.94}_{\pm{5.37}}$ (\textcolor{blue}{$65.06$}) 
  &$63.23_{\pm{17.68}}$ (\textcolor{blue}{$36.77$})    & ${68.02}_{\pm{12.03}}$ (\textcolor{blue}{$31.98$}) 
   &$99.87_{\pm{0.05}}$(\textcolor{blue}{$0.13$})    & $99.60_{\pm{0.08}}$ (\textcolor{blue}{$0.37$}) 
   &$92.80_{\pm{1.28}}$ (\textcolor{blue}{$2.03$})     & ${92.96}_{\pm{0.85}}$ (\textcolor{blue}{$0.03$}) & $1.81$
 \\
 \GA  &$77.51_{\pm{3.47}}$ (\textcolor{blue}{$22.49$})      & ${83.93}_{\pm{2.14}}$ (\textcolor{blue}{$16.07$}) 
  &$80.13_{\pm{4.27}}$ (\textcolor{blue}{$19.87$})    & ${88.04}_{\pm{3.18}}$ (\textcolor{blue}{$11.96$}) 
   &$96.09_{\pm{0.13}}$(\textcolor{blue}{$3.91$})    & ${97.33}_{\pm{0.08}}$ (\textcolor{blue}{$2.64$}) 
   &$88.80_{\pm{1.33}}$ (\textcolor{blue}{$6.03$})     & ${89.95}_{\pm{0.78}}$ (\textcolor{blue}{$2.98$}) & $0.27$
 \\
\IU &$88.58_{\pm{0.86}}$ (\textcolor{blue}{$11.42$})      & ${98.78}_{\pm{0.44}}$ (\textcolor{blue}{$1.22$}) 
  &$92.27_{\pm{1.14}}$ (\textcolor{blue}{$7.73$})    & ${99.91}_{\pm{0.05}}$ (\textcolor{blue}{$0.09$}) 
   &$96.89_{\pm{0.27}}$(\textcolor{blue}{$3.11$})    & ${93.18}_{\pm{0.28}}$ (\textcolor{blue}{$6.79$}) 
   &$89.81_{\pm{1.01}}$ (\textcolor{blue}{$5.02$})     & ${87.45}_{\pm{0.81}}$ (\textcolor{blue}{$5.48$}) & $2.51$
\\

\midrule
\rowcolor{Gray}
\multicolumn{10}{c}{Random data forgetting, VGG-16} \\ 
\midrule
 \retrain &\textcolor{blue}{$7.13_{\pm{0.60}}$}   & \textcolor{blue}{$7.47_{\pm{0.30}}$} 
  &\textcolor{blue}{$13.02_{\pm{0.77}}$} &  \textcolor{blue}{$13.51_{\pm{0.50}}$}
   &\textcolor{blue}{$100.00_{\pm{0.01}}$}  & \textcolor{blue}{$99.93_{\pm{0.01}}$}
   &\textcolor{blue}{$92.80_{\pm{0.17}}$}   & \textcolor{blue}{$91.98_{\pm{0.22}}$} & $30.29$

 \\
 \FT &$0.86_{\pm{0.29}}$ (\textcolor{blue}{$6.27$})      & ${1.46}_{\pm{0.22}}$ (\textcolor{blue}{$6.01$}) 
  &$2.62_{\pm{0.47}}$ (\textcolor{blue}{$10.40$})    & ${3.82}_{\pm{0.41}}$ (\textcolor{blue}{$9.69$}) 
   &$99.76_{\pm{0.12}}$(\textcolor{blue}{$0.24$})    & $99.47_{\pm{0.11}}$ (\textcolor{blue}{$0.53$}) 
   &$92.21_{\pm{0.13}}$ (\textcolor{blue}{$0.59$})     & ${92.03}_{\pm{0.37}}$ (\textcolor{blue}{$0.05$}) & $1.77$
 \\
 \GA & $9.11_{\pm{0.83}}$ (\textcolor{blue}{$1.98$})& $6.91_{\pm{0.96}}$ (\textcolor{blue}{$0.56$})& $7.77_{\pm{1.01}}$ (\textcolor{blue}{$5.25$})& $8.37_{\pm{1.35}}$ (\textcolor{blue}{$5.14$})& $93.08_{\pm{0.93}}$ (\textcolor{blue}{$6.92$})& $93.63_{\pm{1.16}}$ (\textcolor{blue}{$6.30$})& $86.44_{\pm{1.32}}$ (\textcolor{blue}{$6.36$})& $89.22_{\pm{1.59}}$ (\textcolor{blue}{$4.53$}) & 0.31
 \\
\IU &$1.02_{\pm{0.43}}$ (\textcolor{blue}{$6.11$})      & ${3.07}_{\pm{0.50}}$ (\textcolor{blue}{$4.40$}) 
  &$2.51_{\pm{0.61}}$ (\textcolor{blue}{$9.51$})    & ${6.86}_{\pm{0.67}}$ (\textcolor{blue}{$6.65$}) 
   &$99.14_{\pm{0.03}}$(\textcolor{blue}{$0.86$})    & ${97.35}_{\pm{0.31}}$ (\textcolor{blue}{$2.58$}) 
   &$91.01_{\pm{0.29}}$ (\textcolor{blue}{$1.79$})     & ${89.49}_{\pm{0.37}}$ (\textcolor{blue}{$2.49$}) & $2.78$
\\
\midrule
\bottomrule[1pt]
\end{tabular}
}

\end{table*}

To demonstrate the effectiveness of our methods on a larger dataset, we conducted additional experiments on \textbf{ImageNet} \cite{deng2009imagenet} with settings consistent with the class-wise forgetting in Tab.\,\ref{tab: overall_performance}. 
As we can see from Tab.\,\ref{tab: overall_performance_ImageNet}, sparsity reduces the performance gap between exact unlearning (Retrain) and the   approximate unlearning methods (FT and GA). The results are consistent with our  observations in other datasets. Note that the $83\%$ model sparsity (ImageNet, ResNet-18) is used to preserve the TA performance after one-shot magnitude (OMP) pruning.

\begin{table*}[ht]
\centering
\caption{\footnotesize{Performance overview of MU vs. sparsity on ImageNet considering class-wise forgetting. The content format follows Tab.\,\ref{tab: overall_performance}.}}
\label{tab: overall_performance_ImageNet}
\resizebox{0.80\textwidth}{!}{
\begin{tabular}{c|cc|cc|cc|cc|c}
\toprule[1pt]
\midrule
  \multirow{2}{*}{\MU}& \multicolumn{2}{c|}{{\UA}} & \multicolumn{2}{c|}{{\MIAF}}& \multicolumn{2}{c|}{{\RA}} & \multicolumn{2}{c|}{{\TA}}&{\RTE}  \\ 
  & \multicolumn{1}{c|}{{\textsc{Dense}}}  & \multicolumn{1}{c|}{$\mathbf{83\%}$ \textbf{Sparsity}}
    & \multicolumn{1}{c|}{\textsc{Dense}}  & \multicolumn{1}{c|}{$\mathbf{83\%}$ \textbf{Sparsity}}
    & \multicolumn{1}{c|}{\textsc{Dense}}  & \multicolumn{1}{c|}{$\mathbf{83\%}$ \textbf{Sparsity}}
      & \multicolumn{1}{c|}{\textsc{Dense}}  & \multicolumn{1}{c|}{$\mathbf{83\%}$ \textbf{Sparsity}} & (hours)
  \\

\midrule
\rowcolor{Gray}
\multicolumn{10}{c}{Class-wise forgetting, ImageNet} \\
\midrule
 \retrain &\textcolor{blue}{$100.00$}&\textcolor{blue}{$100.00$}&\textcolor{blue}{$100.00$}&\textcolor{blue}{$100.00$}&\textcolor{blue}{$71.75$}&\textcolor{blue}{$69.18$}&\textcolor{blue}{$69.49$}&\textcolor{blue}{$68.86$} & 26.18
\\
 \FT & $63.60$ (\textcolor{blue}{$36.40$})& $74.66$ (\textcolor{blue}{$25.34$})& $68.61$ (\textcolor{blue}{$31.39$})& $81.43$ (\textcolor{blue}{$18.57$})& $72.45$ (\textcolor{blue}{$0.70$})& $69.36$ (\textcolor{blue}{$0.18$})& $69.80$ (\textcolor{blue}{$0.31$})& $68.77$ (\textcolor{blue}{$0.09$}) & 2.87
 \\
 \GA & $85.10$ (\textcolor{blue}{$14.90$})& $90.21$ (\textcolor{blue}{$9.79$})& $87.46$ (\textcolor{blue}{$12.54$})& $94.25$ (\textcolor{blue}{$5.75$})& $65.93$ (\textcolor{blue}{$5.82$})& $62.94$ (\textcolor{blue}{$6.24$})& $64.62$ (\textcolor{blue}{$4.87$})& $64.65$ (\textcolor{blue}{$4.21$}) & 0.01
 \\
\midrule
\bottomrule[1pt]
\end{tabular}
}
\end{table*}

\noindent \textbf{Performance of $\ell_1$ sparsity-aware {\MU} on additional datasets.}
As seen in Fig.\,\ref{fig: results_l1_sparse_unlearn_others},  {\MUSparse} significantly reduces the gap between approximate and exact unlearning methods across various datasets (CIFAR-100 \cite{krizhevsky2009learning}, SVHN \cite{netzer2011reading}, ImageNet \cite{deng2009imagenet}) in different unlearning scenarios. It notably outperforms other methods in {\UA} and {\MIAF} metrics while preserving acceptable {\RA} and {\TA} performances, thus becoming a practical choice for unlearning scenarios. In class-wise and random data forgetting cases, {\MUSparse} exhibits performance on par with {\retrain} in {\UA} and {\MIAF} metrics. Importantly, the use of {\MUSparse} consistently enhances forgetting metrics with an insignificant rise in computational cost compared with {\FT}, underscoring its effectiveness and efficiency in diverse unlearning scenarios. For detailed numerical results, please refer to Tab.\,\ref{tab: sparse_MU vs MU}. 

\begin{figure*}[htb]
\centerline{
\begin{tabular}{ccccc}

\hspace{-1.2mm}
\includegraphics[width=25mm,height=!]{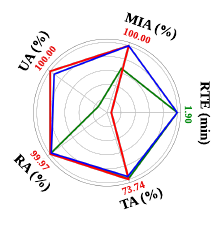}
\hspace{-1.2mm}
 & 
\hspace{-1.2mm}
 \includegraphics[width=25mm,height=!]{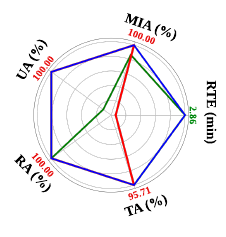}
\hspace{-1.2mm}
& 
\hspace{-1.2mm}
\includegraphics[width=25mm,height=!]{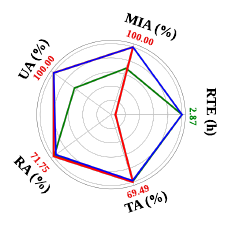}
\hspace{-1.2mm}
 & 
\hspace{-1.2mm}
 \includegraphics[width=25mm,height=!]{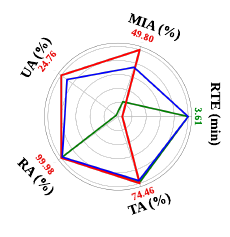}
\hspace{-1.2mm}
 &
\hspace{-1.2mm}
 \includegraphics[width=25mm,height=!]{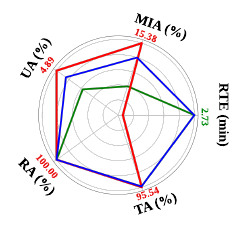}
\hspace{-1.2mm}
\\
\scriptsize{CIFAR-100} & \scriptsize{SVHN} & \scriptsize{ImageNet} & \scriptsize{CIFAR-100} & \scriptsize{SVHN} \\
\scriptsize{Class-wise forgetting} & \scriptsize{Class-wise forgetting} & \scriptsize{Class-wise forgetting} & \scriptsize{Random data forgetting} & \scriptsize{Random data forgetting}
 \\
\multicolumn{5}{c}{\includegraphics[width=50mm,height=!]{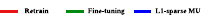}}\\
\end{tabular}}
\vspace*{-1.5mm}
\caption{
Performance   of sparsity-aware unlearning vs. {\FT} and {\retrain} on class-wise forgetting and random data forgetting under ResNet-18 on different datasets. 
 Each metric is normalized to $[0,1]$ based on the best result  across unlearning methods for ease of visualization, while the actual best value is provided. The figure format is consistent with Fig.\,\ref{fig: results_l1_sparse_unlearn}.
}
  \label{fig: results_l1_sparse_unlearn_others}
\vspace*{-5mm}
\end{figure*}

\begin{table*}[htb]
\centering
\caption{Performance   of sparsity-aware {\MU} vs. {\retrain}, {\FT}, {{\GA}} and {\IU} considering class-wise forgetting and random data forgetting, where the table format and setup are consistent with  Tab.\,\ref{tab: overall_performance}. 
The unit of   RTE is minutes for all datasets, except ImageNet. For ImageNet, indicated by an asterisk ($\ast$),  RTE is  measured by hours.
}
\label{tab: sparse_MU vs MU}
\resizebox{0.6\textwidth}{!}{
\begin{tabular}{c|c|c|c|c|c}
\toprule[1pt]
\midrule
  \multirow{1}{*}{\MU}& \multicolumn{1}{c|}{{\UA}} & \multicolumn{1}{c|}{{\MIAF}}& \multicolumn{1}{c|}{{\RA}} & \multicolumn{1}{c|}{{\TA}}&{{\RTE} (min)} 
  \\

\midrule
\rowcolor{Gray}
\multicolumn{6}{c}{Class-wise forgetting, CIFAR-10} \\
\midrule
\retrain & \textcolor{blue}{100.00$_{\pm{0.00}}$} &\textcolor{blue}{100.00$_{\pm{0.00}}$} & \textcolor{blue}{100.00$_{\pm{0.00}}$} & \textcolor{blue}{94.83$_{\pm{0.11}}$} & 43.23
\\
\FT &$22.53_{\pm{8.16}}$ (\textcolor{blue}{$77.47$})&$75.00_{\pm{14.68}}$ (\textcolor{blue}{${25.00}$})
  &$99.87_{\pm{0.04}}$ (\textcolor{blue}{$0.13$}) &$94.31_{\pm{0.19}}$ (\textcolor{blue}{$0.52$})
&   2.52 \\
 \GA &$93.08_{\pm{2.29}}$ (\textcolor{blue}{6.92}) 
& $94.03_{\pm{3.27}}$ (\textcolor{blue}{5.97})
& $92.60_{\pm{0.25}}$ (\textcolor{blue}{$7.40$})
& $86.64_{\pm{0.28}}$ (\textcolor{blue}{$8.19$})
&   0.33
\\
\IU  
&$87.82_{\pm{2.15}} $ (\textcolor{blue}{$12.18$})
 & $95.96_{\pm0.21}$ (\textcolor{blue}{$4.04$})

 &$97.98_{\pm{0.21}}$ (\textcolor{blue}{$2.02$}) 
 
 &$91.42_{\pm{0.21}}$ (\textcolor{blue}{$3.41$})
 & 3.25
\\
\MUSparse 
&$100.00_{\pm{0.00}} $ (\textcolor{blue}{$0.00$})
 & $100.00_{\pm0.00}$ (\textcolor{blue}{$0.00$})

 &$98.99_{\pm{0.12}}$ (\textcolor{blue}{$1.01$}) 
 
 &$93.40_{\pm{0.43}}$ (\textcolor{blue}{$1.43$})
 & 2.53
\\
\midrule
\rowcolor{Gray}
\multicolumn{6}{c}{Class-wise forgetting, CIFAR-100} \\
\midrule
 \retrain &\textcolor{blue}{$100.00_{\pm{0.00}}$}   
  &\textcolor{blue}{$100.00_{\pm{0.00}}$}   
   &\textcolor{blue}{$99.97_{\pm{0.01}}$}   
   &\textcolor{blue}{$73.74_{\pm{0.19}}$}   
& $48.45$
\\
 \FT &$15.00_{\pm{4.86}}$ (\textcolor{blue}{$85.00$})      
  &$66.11_{\pm{4.53}}$ (\textcolor{blue}{$33.89$})    
   &$99.87_{\pm{0.05}}$(\textcolor{blue}{$0.10$})   
   &$74.88_{\pm{0.18}}$ (\textcolor{blue}{$1.14$})  &$1.90$
\\
  \GA  &$81.47_{\pm{0.32}}$(\textcolor{blue}{$18.53$})  
 & $93.47_{\pm{4.56}}$ (\textcolor{blue}{$6.53$})  
 & $90.33_{\pm{1.71}}$ (\textcolor{blue}{$9.64$}) 
 & $64.94_{\pm{0.74}}$ (\textcolor{blue}{$8.80$})  
& $0.21$
 \\
\IU &$84.12_{\pm{0.34}}$ (\textcolor{blue}{$15.88$}) 
& $98.44_{\pm{0.45}}$ (\textcolor{blue}{$1.56$}) 
& $96.23_{\pm{0.02}}$ (\textcolor{blue}{$3.74$})  
&$71.24_{\pm{0.22}}$ (\textcolor{blue}{$2.50$}) 
& $4.30$
\\
\MUSparse 
& $93.11_{\pm{0.49}}$ (\textcolor{blue}{$6.89$}) 
& $100.00_{\pm{0.00}}$ (\textcolor{blue}{$0.00$}) 
& $98.00_{\pm{0.07}}$ (\textcolor{blue}{$1.97$})  
& $70.68_{\pm{0.26}}$ (\textcolor{blue}{$3.06$}) 
& $1.91$
\\
\midrule
\rowcolor{Gray}
\multicolumn{6}{c}{Class-wise forgetting, SVHN} \\
\midrule
\retrain &\textcolor{blue}{$100.00_{\pm{0.00}}$}  
 & \textcolor{blue}{$100.00_{\pm{0.00}}$}
& \textcolor{blue}{$100.00_{\pm{0.00}}$} 
 &  \textcolor{blue}{$95.71_{\pm{0.12}}$} & $42.84$
 \\
 \FT & {$11.48_{\pm{8.12}}$ } (\textcolor{blue}{$88.52$})    
& $86.12_{\pm{9.62}}$ (\textcolor{blue}{$13.88$}) 
& $100.00_{\pm{0.00}}$ (\textcolor{blue}{$0.00$}) 
& $95.99_{\pm{0.07}}$ (\textcolor{blue}{$0.28$}) & $2.86$
 \\
    \GA &$83.87_{\pm{0.19}}$ (\textcolor{blue}{$16.13$})  
  & $99.97_{\pm{0.02}}$ (\textcolor{blue}{$0.03$})
  & $99.60_{\pm{0.15}}$ (\textcolor{blue}{$0.40$}) 
  &  $95.27_{\pm{0.02}}$ (\textcolor{blue}{$0.44$}) & $0.28$
  \\
\IU &$95.11_{\pm{0.02}}$ (\textcolor{blue}{$4.89$})   
&$99.89_{\pm{0.04}}$ (\textcolor{blue}{$0.11$}) 
&$100.00_{\pm{0.00}}$ (\textcolor{blue}{$0.00$})   
&$95.70_{\pm{0.09}}$ (\textcolor{blue}{$0.01$})   & $3.19$  
\\
\MUSparse &$100.00_{\pm{0.00}}$ (\textcolor{blue}{$0.00$})   
&$100.00_{\pm{0.00}}$ (\textcolor{blue}{$0.00$}) 
&$99.99_{\pm{0.01}}$ (\textcolor{blue}{$0.01$})   
&$95.88_{\pm{0.14}}$ (\textcolor{blue}{$0.17$})   & $2.88$
\\
\midrule
\rowcolor{Gray}
\multicolumn{6}{c}{Class-wise forgetting, ImageNet} \\
\midrule
 \retrain &\textcolor{blue}{$100.00_{\pm{0.00}}$}&\textcolor{blue}{$100.00_{\pm{0.00}}$}&\textcolor{blue}{$71.75_{\pm{0.45}}$}&\textcolor{blue}{$69.49_{\pm{0.27}}$} & $26.18^\ast$
\\
 \FT & $63.60_{\pm{7.11}}$ (\textcolor{blue}{$36.40$})& $68.61_{\pm{9.04}}$ (\textcolor{blue}{$31.39$})& $72.45_{\pm{0.16}}$ (\textcolor{blue}{$0.70$})& $69.80_{\pm{0.23}}$ (\textcolor{blue}{$0.31$}) & $2.87^\ast$
\\
 \GA & $85.10_{\pm{5.92}}$ (\textcolor{blue}{$14.90$})& $87.46_{\pm{7.20}}$ (\textcolor{blue}{$12.54$})& $65.93_{\pm{0.49}}$ (\textcolor{blue}{$5.82$})& $64.62_{\pm{0.82}}$ (\textcolor{blue}{$4.87$}) & $0.01^\ast$ \\
\IU  & $43.35_{\pm{5.26}}$ (\textcolor{blue}{$56.65$})& $60.83_{\pm{6.17}}$ (\textcolor{blue}{$39.17$})& $66.28_{\pm{0.77}}$ (\textcolor{blue}{$5.47$})& $66.25_{\pm{0.53}}$ (\textcolor{blue}{$3.24$}) & $3.14^\ast$
\\
\MUSparse & $99.85_{\pm{0.07}}$ (\textcolor{blue}{$0.15$})& $100.00_{\pm{0.00}}$ (\textcolor{blue}{$0.00$})& $68.07_{\pm{0.13}}$ (\textcolor{blue}{$3.68$})& $68.01_{\pm{0.21}}$ (\textcolor{blue}{$1.48$}) & $2.87^\ast$
\\
\midrule
\rowcolor{Gray}
\multicolumn{6}{c}{Random data forgetting, CIFAR-10} \\
\midrule
 \retrain &\textcolor{blue}{$5.41_{\pm{0.11}}$}&\textcolor{blue}{$13.12_{\pm{0.14}}$}&\textcolor{blue}{$100.00_{\pm{0.00}}$}&\textcolor{blue}{$94.42_{\pm{0.09}}$}& 42.15 
\\
 \FT & $6.83_{\pm{0.51}}$ (\textcolor{blue}{$1.42$})& $14.97_{\pm{0.62}}$ (\textcolor{blue}{$1.85$})& $96.61_{\pm{0.25}}$ (\textcolor{blue}{$3.39$})& $90.13_{\pm{0.26}}$ (\textcolor{blue}{$4.29$}) & 2.33  
 \\
  \GA & $7.54_{\pm{0.29}}$ (\textcolor{blue}{$2.13$})& $10.04_{\pm{0.31}}$ (\textcolor{blue}{$3.08$})& $93.31_{\pm{0.04}}$ (\textcolor{blue}{$6.69$})& $89.28_{\pm{0.07}}$ (\textcolor{blue}{$5.14$})& 0.31
 \\
  \IU & $2.03_{\pm{0.43}}$ (\textcolor{blue}{$3.38$})& $5.07_{\pm{0.74}}$ (\textcolor{blue}{$8.05$})& $98.26_{\pm{0.29}}$ (\textcolor{blue}{$1.74$})& $91.33_{\pm{0.22}}$ (\textcolor{blue}{$3.09$}) & 3.22  
\\
\MUSparse & $5.35_{\pm{0.22}}$ (\textcolor{blue}{$0.06$})& $12.71_{\pm{0.31}}$ (\textcolor{blue}{$0.41$})& $97.39_{\pm{0.19}}$ (\textcolor{blue}{$2.61$})& $91.26_{\pm{0.20}}$ (\textcolor{blue}{$3.16$}) & 2.34 
\\
\midrule
\rowcolor{Gray}
\multicolumn{6}{c}{Random data forgetting, CIFAR-100} \\
\midrule
 \retrain &\textcolor{blue}{$24.76_{\pm{0.12}}$}&\textcolor{blue}{$49.80_{\pm{0.26}}$}&\textcolor{blue}{$99.98_{\pm{0.02}}$}&\textcolor{blue}{$74.46_{\pm{0.08}}$}& 48.70 
\\
 \FT & $0.78_{\pm{0.34}}$ (\textcolor{blue}{$23.98$})& $11.13_{\pm{0.40}}$ (\textcolor{blue}{$38.67$})& $99.93_{\pm{0.02}}$ (\textcolor{blue}{$0.05$})& $75.14_{\pm{0.09}}$ (\textcolor{blue}{$0.68$}) & 3.74  
 \\
  \GA  &$0.04_{\pm{0.02}}$(\textcolor{blue}{$24.72$})  
 & $3.80_{\pm{0.87}}$ (\textcolor{blue}{$46.00$})   
 & $99.97_{\pm{0.01}}$ (\textcolor{blue}{$0.01$}) 
 & $74.07_{\pm{0.11}}$ (\textcolor{blue}{$0.39$})  
& $0.24$
 \\
  \IU & $ 1.53_{\pm{0.36}}$ (\textcolor{blue}{$23.23$})& $6.58_{\pm{0.42}}$ (\textcolor{blue}{$43.22$})& $99.01_{\pm{0.28}}$ (\textcolor{blue}{$0.97$})& $71.76_{\pm{0.31}}$ (\textcolor{blue}{$2.70$}) & 3.80  	
\\
\MUSparse & $20.77_{\pm{0.27}}$ (\textcolor{blue}{$3.99$})& $36.80_{\pm{0.44}}$ (\textcolor{blue}{$13.00$})& $98.26_{\pm{0.15}}$ (\textcolor{blue}{$1.72$})& $71.52_{\pm{0.21}}$ (\textcolor{blue}{$2.94$}) & 3.76 
\\

\midrule
\rowcolor{Gray}
\multicolumn{6}{c}{Random data forgetting, SVHN} \\
\midrule
 \retrain &\textcolor{blue}{$4.89_{\pm{0.11}}$}&\textcolor{blue}{$15.38_{\pm{0.14}}$}&\textcolor{blue}{$100.00_{\pm{0.00}}$}&\textcolor{blue}{$95.54_{\pm{0.09}}$} & 42.71
\\
 \FT & $2.28_{\pm{1.41}}$ (\textcolor{blue}{$2.61$})& $6.14_{\pm{3.30}}$ (\textcolor{blue}{$9.24$})& $99.71_{\pm{0.33}}$ (\textcolor{blue}{$0.29$})& $94.77_{\pm{0.87}}$ (\textcolor{blue}{$0.77$}) & 2.73
 \\
 \GA & $0.99_{\pm{0.42}}$ (\textcolor{blue}{$3.90$})& $3.07_{\pm{0.53}}$ (\textcolor{blue}{$12.31$})& $99.43_{\pm{0.22}}$ (\textcolor{blue}{$0.57$})& $94.03_{\pm{0.21}}$ (\textcolor{blue}{$1.51$}) & 0.26 \\
\IU &$3.48_{\pm{0.13}}$ (\textcolor{blue}{$1.41$})   
&$9.44_{\pm{0.27}}$ (\textcolor{blue}{$5.94$}) 
&$96.30_{\pm{0.08}}$ (\textcolor{blue}{$3.70$})   
&$91.59_{\pm{0.11}}$ (\textcolor{blue}{$3.95$})    & $3.21$
\\
\MUSparse & $4.06_{\pm{0.14}}$ (\textcolor{blue}{$0.83$})& $11.80_{\pm{0.22}}$ (\textcolor{blue}{$3.58$})& $99.96_{\pm{0.01}}$ (\textcolor{blue}{$0.04$})& $94.98_{\pm{0.03}}$ (\textcolor{blue}{$0.56$}) & 2.73 
\\
			
\midrule
\bottomrule[1pt]
\end{tabular}
}
\vspace*{-3mm}

\end{table*}

\begin{table}[htb]
\centering
\caption{Performance   of  {\MUSparse} vs. {\retrain} and {\FT} on (\textbf{Swin Transformer}, CIFAR-10).}
\label{tab: vit}
\resizebox{0.49\textwidth}{!}{
\begin{tabular}{c|c|c|c|c|c}
\toprule[1pt]
\midrule
  \multirow{1}{*}{\MU}& \multicolumn{1}{c|}{{\UA}} & \multicolumn{1}{c|}{{\MIAF}}& \multicolumn{1}{c|}{{\RA}} & \multicolumn{1}{c|}{{\TA}}&{{\RTE} (min)} 
  \\
\midrule
\rowcolor{Gray}
\multicolumn{6}{c}{Class-wise forgetting} \\
\midrule
\retrain & \textcolor{blue}{100.00} &\textcolor{blue}{100.00} & \textcolor{blue}{100.00} & \textcolor{blue}{80.14} & 153.60
\\
\FT &$8.56$ (\textcolor{blue}{$91.44$})&$22.46$ (\textcolor{blue}{${77.54}$})
  &$99.92$ (\textcolor{blue}{$0.08$}) &$79.72$ (\textcolor{blue}{$0.42$})
&   3.87
\\
\textbf{\MUSparse} 
&$98.80 $ (\textcolor{blue}{$1.20$})
 & $100.00$ (\textcolor{blue}{$0.00$})

 &$98.25$ (\textcolor{blue}{$1.75$}) 
 
 &$80.20$ (\textcolor{blue}{$0.06$})
 & 3.89
\\
\midrule
\rowcolor{Gray}
\multicolumn{6}{c}{Random data forgetting} \\
\midrule		
 \retrain &\textcolor{blue}{$21.48$}&\textcolor{blue}{$28.44$}&\textcolor{blue}{$100.00$}&\textcolor{blue}{$78.59$}& 155.06 
\\ 
 \FT & $0.16$ (\textcolor{blue}{$21.32$})& $1.26$ (\textcolor{blue}{$27.18$})& $99.80$ (\textcolor{blue}{$0.20$})& $79.54$ (\textcolor{blue}{$0.95$}) & 7.77    
\\ 			
\textbf{\MUSparse}  & $9.22$ (\textcolor{blue}{$12.26$})& $18.33$ (\textcolor{blue}{$10.11$})& $97.92$ (\textcolor{blue}{$2.08$})& $79.09$ (\textcolor{blue}{$0.50$}) & 7.84 
\\
\midrule
\bottomrule[1pt]
\end{tabular}
}
\vspace*{-3mm}
\end{table}
{\noindent \textbf{Performance of $\ell_1$ sparsity-aware {\MU} on additional architectures.}}
Tab.\,\ref{tab: vit} presents an additional application to Swin Transformer on  CIFAR-10. To facilitate a comparison between approximate unlearning methods (including the {\FT} baseline and the proposed $\ell_1$-sparse MU) and Retrain, we train the transformer from scratch on CIFAR-10. This could potentially decrease testing accuracy compared with fine-tuning on a pre-trained model over a larger, pre-trained dataset. As we can see,  our proposed $\ell_1$-sparse MU  leads to a much smaller performance gap with {\retrain} compared to {\FT}. In particular, class-wise forgetting exhibited a remarkable $90.24\%$ increase in UA, accompanied by a slight reduction in RA.

\noindent \textbf{Performance of `prune first, then unlearn’ and $\ell_1$ sparsity-aware {\MU} on different model sizes.}
Further, Tab.\,\ref{tab: overall_perfoamnce_arch_20} and Tab.\,\ref{tab: overall_perfoamnce_arch_50} present the unlearning performance versus different model sizes in the ResNet family, involving both ResNet-20s and ResNet-50 on CIFAR-10, in addition to ResNet-18 in Tab.\,\ref{tab: overall_performance}. As we can see, sparsity consistently diminishes the unlearning gap with Retrain (indicated by highlighted numbers, with smaller values being preferable). It's worth noting that while both ResNet-20s and ResNet-50 benefit from sparsity, the suggested sparsity ratio is 90\% for ResNet-20s and slightly lower than 95\% for ResNet-50 when striking the balance between MU and generalization. 

\begin{table*}[htb]
\centering

\caption{\footnotesize{MU performance on (\textbf{ResNet-20s}, CIFAR-10) using   `prune first, then unlearn'  (applying to the OMP-resulted 90\% sparse model)  and `sparse-aware unlearning' (applying to the original dense model). The   performance is reported in the form $a_{\pm b}$, with mean $a$ and standard deviation $b$ computed over $10$ independent trials.  
A performance gap  against \textcolor{blue}{{\retrain}} is provided 
in (\textcolor{blue}{$\bullet$}).Smaller performance gap from Retrain is better in the context of machine unlearning.
} }
\vspace*{-1mm}
\label{tab: overall_perfoamnce_arch_20}
\resizebox{0.95\textwidth}{!}{
\begin{tabular}{c|cc|cc|cc|cc|c}
\toprule[1pt]
\midrule
  \multirow{2}{*}{\MU}& \multicolumn{2}{c|}{{\UA}} & \multicolumn{2}{c|}{{\MIAF}}& \multicolumn{2}{c|}{{\RA}} & \multicolumn{2}{c|}{{\TA}}&{\RTE}  \\ 
  & \multicolumn{1}{c|}{{\textsc{Dense}}}  & \multicolumn{1}{c|}{\textbf{$90\%$ Sparsity}}
    & \multicolumn{1}{c|}{\textsc{Dense}}  & \multicolumn{1}{c|}{\textbf{$90\%$ Sparsity}}
    & \multicolumn{1}{c|}{\textsc{Dense}}  & \multicolumn{1}{c|}{\textbf{$90\%$ Sparsity}}
      & \multicolumn{1}{c|}{\textsc{Dense}}  & \multicolumn{1}{c|}{\textbf{$90\%$ Sparsity}} & (min)
  \\

\midrule
\rowcolor{Gray}
\multicolumn{10}{c}{Class-wise forgetting} \\
\midrule
 \retrain &\textcolor{blue}{$100.00_{\pm{0.00}}$}   & \textcolor{blue}{$100.00_{\pm{0.00}}$} 
  &\textcolor{blue}{$100.00_{\pm{0.00}}$} &  \textcolor{blue}{$100.00_{\pm{0.00}}$}
   &\textcolor{blue}{$99.76_{\pm{0.03}}$}  & \textcolor{blue}{$92.95_{\pm{0.20}}$}
   &\textcolor{blue}{$92.22_{\pm{0.20}}$}   & \textcolor{blue}{$88.58_{\pm{0.29}}$} & $25.27$

 \\
 \FT &$83.10_{\pm{4.83}}$ (\textcolor{blue}{$16.90$})      & ${91.67}_{\pm{3.81}}$ (\textcolor{blue}{$8.33$}) 
  &$97.17_{\pm{0.75}}$ (\textcolor{blue}{$2.83$})    & ${99.37}_{\pm{0.29}}$ (\textcolor{blue}{$0.63$}) 
   &$98.14_{\pm{0.28}}$(\textcolor{blue}{$1.62$})    & $93.33_{\pm{0.80}}$ (\textcolor{blue}{$0.38$}) 
   &$90.99_{\pm{0.40}}$ (\textcolor{blue}{$1.23$})     & ${88.90}_{\pm{0.63}}$ (\textcolor{blue}{$0.32$}) & $1.57$
 \\
 \GA  &$88.48_{\pm{3.47}}$ (\textcolor{blue}{$11.52$})      & ${90.57}_{\pm{2.14}}$ (\textcolor{blue}{$9.43$}) 
  &$92.55_{\pm{4.27}}$ (\textcolor{blue}{$7.45$})    & ${97.37}_{\pm{2.18}}$ (\textcolor{blue}{$2.63$}) 
   &$91.42_{\pm{0.53}}$(\textcolor{blue}{$8.34$})    & ${86.75}_{\pm{0.88}}$ (\textcolor{blue}{$6.20$}) 
   &$85.46_{\pm{1.33}}$ (\textcolor{blue}{$6.76$})     & ${83.33}_{\pm{0.78}}$ (\textcolor{blue}{$5.25$}) & $0.10$
 \\
   \textbf{\MUSparse} &$98.57_{\pm{0.86}}$ (\textcolor{blue}{$1.43$}) & n/a
  & $100.00_{\pm{0.00}}$ (\textcolor{blue}{$0.00$})  & n/a
  & $96.18_{\pm{0.91}}$ (\textcolor{blue}{$3.58$})  & n/a  
  &  $90.18_{\pm{0.14}}$ (\textcolor{blue}{$2.04$})    & n/a
  & 1.60
  \\
\midrule
\rowcolor{Gray}
\multicolumn{10}{c}{Random data forgetting} \\
\midrule
 \retrain &\textcolor{blue}{$8.02_{\pm{0.36}}$}   & \textcolor{blue}{$12.33_{\pm{0.38}}$} 
 & \textcolor{blue}{$14.94_{\pm{0.46}}$}  &  \textcolor{blue}{$16.46_{\pm{0.83}}$ }
& \textcolor{blue}{$100.00_{\pm{0.00}}$}   &  \textcolor{blue}{$92.33_{\pm{0.18}}$}  
 &  \textcolor{blue}{$91.10_{\pm{0.27}}$} & \textcolor{blue}{$86.46_{\pm{0.02}}$ } &$25.29$
 \\
 \FT & {$3.46_{\pm{0.32}}$ } (\textcolor{blue}{$4.56$})    & ${8.93}_{\pm{0.52}}$ (\textcolor{blue}{$3.40$}) 
& $9.33_{\pm{0.45}}$ (\textcolor{blue}{$5.61$})   &${12.62}_{\pm{0.51}}$  (\textcolor{blue}{$3.84$}) 
& $98.57_{\pm{0.20}}$ (\textcolor{blue}{$1.43$})    &  $93.59_{\pm{0.33}}$ (\textcolor{blue}{$1.26$}) 
& $90.71_{\pm{0.14}}$ (\textcolor{blue}{$0.39$})  &  $88.15_{\pm{0.12}}$ (\textcolor{blue}{$1.69$}) & $1.58$
 \\
  \GA &$1.84_{\pm{0.53}}$ (\textcolor{blue}{$6.18$})    & ${6.88}_{\pm{0.41}}$ (\textcolor{blue}{$5.45$}) 
  & $6.53_{\pm{0.42}}$ (\textcolor{blue}{$8.41$})  &  ${9.57}_{\pm{0.56}}$ (\textcolor{blue}{$6.89$}) 
  & $97.41_{\pm{0.21}}$ (\textcolor{blue}{$2.59$})  &$94.78_{\pm{0.11}}$ (\textcolor{blue}{$2.45$})  
  &  $91.03_{\pm{0.74}}$ (\textcolor{blue}{$0.07$})    & $89.15_{\pm{0.31}}$ (\textcolor{blue}{$2.69$}) 
  & $0.10$
  \\
   \textbf{\MUSparse} &$6.44_{\pm{0.23}}$ (\textcolor{blue}{$1.58$}) & n/a
  & $13.15_{\pm{0.31}}$ (\textcolor{blue}{$1.79$})  & n/a
  & $96.31_{\pm{0.14}}$ (\textcolor{blue}{$3.69$})  & n/a  
  &  $89.14_{\pm{0.26}}$ (\textcolor{blue}{$1.96$})    & n/a
  & 1.58
\\
\midrule
\bottomrule[1pt]
\end{tabular}
}
\vspace*{-2mm}
\end{table*}

\begin{table*}[htb!]
\centering
\caption{\footnotesize{MU performance on (\textbf{ResNet-50}, CIFAR-10) 
following the format of Table\,\ref{tab: overall_perfoamnce_arch_20}.)
} }
\vspace*{-1mm}
\label{tab: overall_perfoamnce_arch_50}
\resizebox{0.95\textwidth}{!}{
\begin{tabular}{c|cc|cc|cc|cc|c}
\toprule[1pt]
\midrule
  \multirow{2}{*}{\MU}& \multicolumn{2}{c|}{{\UA}} & \multicolumn{2}{c|}{{\MIAF}}& \multicolumn{2}{c|}{{\RA}} & \multicolumn{2}{c|}{{\TA}}&{\RTE}  \\ 
  & \multicolumn{1}{c|}{{\textsc{Dense}}}  & \multicolumn{1}{c|}{\textbf{$95\%$ Sparsity}}
    & \multicolumn{1}{c|}{\textsc{Dense}}  & \multicolumn{1}{c|}{\textbf{$95\%$ Sparsity}}
    & \multicolumn{1}{c|}{\textsc{Dense}}  & \multicolumn{1}{c|}{\textbf{$95\%$ Sparsity}}
      & \multicolumn{1}{c|}{\textsc{Dense}}  & \multicolumn{1}{c|}{\textbf{$95\%$ Sparsity}} & (min)
  \\

\midrule
\rowcolor{Gray}
\multicolumn{10}{c}{Class-wise forgetting} \\
\midrule
 \retrain &\textcolor{blue}{$100.00_{\pm{0.00}}$}   & \textcolor{blue}{$100.00_{\pm{0.00}}$} 
  &\textcolor{blue}{$100.00_{\pm{0.00}}$} &  \textcolor{blue}{$100.00_{\pm{0.00}}$}
   &\textcolor{blue}{$100.00_{\pm{0.03}}$}  & \textcolor{blue}{$100.00_{\pm{0.00}}$}
   &\textcolor{blue}{$94.18_{\pm{0.38}}$}   & \textcolor{blue}{$94.12_{\pm{0.07}}$} & $96.29$

 \\
 \FT &$49.76_{\pm{5.04}}$ (\textcolor{blue}{$50.24$})      & ${57.84}_{\pm{3.10}}$ (\textcolor{blue}{$42.16$}) 
  &$84.67_{\pm{6.90}}$ (\textcolor{blue}{$15.33$})    & ${88.20}_{\pm{3.70}}$ (\textcolor{blue}{$11.80$}) 
   &$99.62_{\pm{0.12}}$(\textcolor{blue}{$0.38$})    & $99.65_{\pm{0.06}}$ (\textcolor{blue}{$0.35$}) 
   &$94.11_{\pm{0.30}}$ (\textcolor{blue}{$0.07$})     & ${93.54}_{\pm{0.15}}$ (\textcolor{blue}{$0.58$}) & $6.02$
 \\
 \GA  &$93.41_{\pm{0.24}}$ (\textcolor{blue}{$6.59$})      & ${93.90}_{\pm{0.21}}$ (\textcolor{blue}{$6.10$}) 
  &$95.90_{\pm{0.18}}$ (\textcolor{blue}{$4.10$})    & ${96.22}_{\pm{0.24}}$ (\textcolor{blue}{$3.78$}) 
   &$93.44_{\pm{0.53}}$(\textcolor{blue}{$6.56$})    & ${93.05}_{\pm{0.26}}$ (\textcolor{blue}{$6.95$}) 
   &$87.37_{\pm{0.15}}$ (\textcolor{blue}{$6.81$})     & ${87.22}_{\pm{0.08}}$ (\textcolor{blue}{$6.90$}) & $0.30$
 \\
   \MUSparse &$96.46_{\pm{0.51}}$ (\textcolor{blue}{$3.54$}) & n/a
  & $100.00_{\pm{0.00}}$ (\textcolor{blue}{$0.00$})  & n/a
  & $99.11_{\pm{0.42}}$ (\textcolor{blue}{$0.89$})  & n/a  
  &  $92.83_{\pm{0.10}}$ (\textcolor{blue}{$1.35$})    & n/a
  & 6.05
  \\
\midrule
\rowcolor{Gray}
\multicolumn{10}{c}{Random data forgetting} \\
\midrule
 \retrain &\textcolor{blue}{$5.81_{\pm{0.29}}$}   & \textcolor{blue}{$6.09_{\pm{0.45}}$} 
 & \textcolor{blue}{$11.99_{\pm{0.94}}$}  &  \textcolor{blue}{$12.76_{\pm{0.86}}$ }
& \textcolor{blue}{$100.00_{\pm{0.00}}$}   &  \textcolor{blue}{$99.00_{\pm{0.00}}$}  
 &  \textcolor{blue}{$93.62_{\pm{0.21}}$} & \textcolor{blue}{$93.76_{\pm{0.03}}$ } &$96.30$
 \\
 \FT & {$5.17_{\pm{0.75}}$ } (\textcolor{blue}{$0.64$})    & ${5.84}_{\pm{0.45}}$ (\textcolor{blue}{$0.25$}) 
& $10.93_{\pm{0.94}}$ (\textcolor{blue}{$1.06$})   &${12.17}_{\pm{0.82}}$  (\textcolor{blue}{$0.59$}) 
& $97.64_{\pm{0.22}}$ (\textcolor{blue}{$2.36$})    &  $97.24   _{\pm{0.35}}$ (\textcolor{blue}{$1.76$}) 
& $91.13_{\pm{0.14}}$ (\textcolor{blue}{$2.49$})  &  $90.81_{\pm{0.12}}$ (\textcolor{blue}{$2.95$}) & $6.02$
 \\
  \GA &$3.42_{\pm{0.25}}$ (\textcolor{blue}{$2.39$})    & ${5.77}_{\pm{0.37}}$ (\textcolor{blue}{$0.32$}) 
  & $5.20_{\pm{0.42}}$ (\textcolor{blue}{$6.79$})  &  ${8.73}_{\pm{0.56}}$ (\textcolor{blue}{$4.03$}) 
  & $96.20_{\pm{0.19}}$ (\textcolor{blue}{$3.80$})  &$95.41_{\pm{0.14}}$ (\textcolor{blue}{$3.59$})  
  &  $90.12_{\pm{0.21}}$ (\textcolor{blue}{$3.50$})    & $89.47_{\pm{0.26}}$ (\textcolor{blue}{$4.29$}) 
  & $0.32$
  \\
   \MUSparse &$6.13_{\pm{0.17}}$ (\textcolor{blue}{$0.32$}) & n/a
  & $12.29_{\pm{0.20}}$ (\textcolor{blue}{$0.30$})  & n/a
  & $97.12_{\pm{0.16}}$ (\textcolor{blue}{$2.88$})  & n/a  
  &  $91.12_{\pm{0.15}}$ (\textcolor{blue}{$2.50$})    & n/a
  & 6.10
\\
\midrule
\bottomrule[1pt]
\end{tabular}
}
\vspace*{-3mm}
\end{table*}

\section{Broader Impacts and Limitations}
\label{app: broader_impact}
\noindent \textbf{Broader impacts.} 
Our study on model sparsity-inspired {\MU} provides a versatile solution to forget arbitrary data points and could give a general solution for dealing with different concerns, such as the model's privacy, efficiency, and robustness. 
Moreover, the applicability of our method extends beyond these aspects, with potential impacts in the following areas.
 \ding{172} \textit{Regulatory compliance:} Our method enables industries, such as healthcare and finance, to adhere to regulations that require the forgetting of data after a specified period. This capability ensures compliance while preserving the utility and performance of machine learning models. 
 \ding{173} \textit{Fairness:} 
 Our method could also play a crucial role in addressing fairness concerns by facilitating the unlearning of biased datasets or subsets. By removing biased information from the training data, our method contributes to mitigating bias in machine learning models, ultimately fostering the development of fairer models. 
\ding{174} \textit{ML with adaptation and sustainability:} Our method could promote the dynamic adaptation of machine learning models by enabling the unlearning of outdated information, and thus, could enhance the accuracy and relevance of the models to the evolving trends and dynamics of the target domain. This capability fosters sustainability by ensuring that ML models remain up-to-date and adaptable, thus enabling their continued usefulness and effectiveness over time.

\noindent \textbf{Limitations.} 
One potential limitation of our study is the absence of provable guarantees for {\MUSparse}. Since model sparsification is integrated with model training as a soft regularization, the lack of formal proof may raise concerns about the reliability and robustness of the approach.
Furthermore, while our proposed unlearning framework is generic, its applications have mainly focused on solving computer vision tasks. As a result, its effectiveness in the domain of natural language processing (NLP) remains unverified. This consideration becomes particularly relevant when considering large language models. Therefore, further investigation is necessary for future studies to explore the applicability and performance of the framework in NLP tasks.

\end{document}